\documentclass[sigconf,all]{acmart}
\usepackage{algorithm}
\usepackage{algorithmic}
\usepackage[utf8]{inputenc} % allow utf-8 input
\usepackage[T1]{fontenc}    % use 8-bit T1 fonts      % hyperlinks
\usepackage{url}            % simple URL typesetting
\usepackage{booktabs}       % professional-quality tables
\usepackage{amsfonts}       % blackboard math symbols
\usepackage{nicefrac}       % compact symbols for 1/2, etc.
\usepackage{microtype}      % microtypography
\usepackage{xcolor}         % colors
\usepackage{xspace}
\usepackage{amsmath}
\usepackage{multirow}
\usepackage{rotating}
\usepackage{tabularx}
\usepackage{gensymb}
\usepackage{graphicx}
\usepackage{float}
\usepackage{floatflt}
\usepackage{makecell}
\usepackage{colortbl}
\usepackage{calc}
\usepackage{subcaption}
\usepackage{booktabs}
\usepackage{arydshln}

\makeatletter
\definecolor{colorFst}{HTML}{F59194}      % first
\definecolor{colorSnd}{HTML}{FAC791} % second
\definecolor{colorThd}{HTML}{FFFF99}   % third
\definecolor{url_color}{RGB}{42, 83, 163}
\newcommand{\fs}[1]{\colorbox{colorFst}{\textbf{#1}}}
\newcommand{\nd}[1]{\colorbox{colorSnd}{\textbf{#1}}}     % second
\newcommand{\third}[1]{\colorbox{colorThd}{\textbf{#1}}}

\DeclareRobustCommand\onedot{\futurelet\@let@token\@onedot}
\def\@onedot{\ifx\@let@token.\else.\null\fi\xspace}

\def\eg{\emph{e.g}\onedot} 
\def\ie{\emph{i.e}\onedot}

\def\wrt{w.r.t\onedot} 

\makeatother

\AtBeginDocument{%
  \providecommand\BibTeX{{%
    \normalfont B\kern-0.5em{\scshape i\kern-0.25em b}\kern-0.8em\TeX}}}

\copyrightyear{2024}
\acmYear{2024}
\setcopyright{acmlicensed}
\acmConference[SIGGRAPH Conference Papers '24]{Special Interest Group on Computer Graphics and Interactive Techniques Conference Conference Papers '24}{July 27-August 1, 2024}{Denver, CO, USA}
\acmBooktitle{Special Interest Group on Computer Graphics and Interactive Techniques Conference Conference Papers '24 (SIGGRAPH Conference Papers '24), July 27-August 1, 2024, Denver, CO, USA}
\acmDOI{10.1145/3641519.3657417}
\acmISBN{979-8-4007-0525-0/24/07}

\begin{document}

\title[GaussianPrediction]{GaussianPrediction: Dynamic 3D Gaussian Prediction for Motion Extrapolation and Free View Synthesis}

\acmSubmissionID{341}

\author{Boming Zhao}
\authornote{Boming Zhao and Yuan Li contributed equally to this work.}
\email{bmzhao@zju.edu.cn}
\affiliation{%
  \institution{Zhejiang University}
  \city{Hangzhou}
  \country{China}
}

\author{Yuan Li}
\authornotemark[1]
\email{yuan_li@zju.edu.cn}
\affiliation{%
  \institution{Zhejiang University}
  \city{Hangzhou}
  \country{China}
}

\author{Ziyu Sun}
\email{sunzy2121@mails.jlu.edu.cn}
\affiliation{%
  \institution{Jilin University}
  \city{Changchun}
  \country{China}
}

\author{Lin Zeng}
\email{22251265@zju.edu.cn}
\affiliation{%
  \institution{Zhejiang University}
  \city{Hangzhou}
  \country{China}
}

\author{Yujun Shen}
\email{shenyujun0302@gmail.com}
\affiliation{%
  \institution{Ant Group}
  \city{Hangzhou}
  \country{China}
}

\author{Rui Ma}
\email{ruim@jlu.edu.cn}
\affiliation{%
  \institution{Jilin University}
  \city{Changchun}
  \country{China}
}

\author{Yinda Zhang}
\email{yindaz@google.com}
\affiliation{%
  \institution{Google Inc.}
  \city{Mountain View}
  \country{USA}
}

\author{Hujun Bao}
\email{bao@cad.zju.edu.cn}
\affiliation{%
  \institution{Zhejiang University}
  \city{Hangzhou}
  \country{China}
}

\author{Zhaopeng Cui}
\email{zhpcui@gmail.com}
\authornote{Corresponding author.}
\affiliation{%
  \institution{Zhejiang University}
  \city{Hangzhou}
  \country{China}
}


\renewcommand{\shortauthors}{Zhao et al.}

\begin{abstract}
  Forecasting future scenarios in dynamic environments is essential for intelligent decision-making and navigation, a challenge yet to be fully realized in computer vision and robotics. 
  Traditional approaches like video prediction and novel-view synthesis either lack the ability to forecast from arbitrary viewpoints or to predict temporal dynamics. 
  In this paper, we introduce GaussianPrediction, a novel framework that empowers 3D Gaussian representations with dynamic scene modeling and future scenario synthesis in dynamic environments.
  GaussianPrediction can forecast future states from any viewpoint, using video observations of dynamic scenes. To this end, we first propose a 3D Gaussian canonical space with deformation modeling to capture the appearance and geometry of dynamic scenes, and integrate the lifecycle property into Gaussians for irreversible deformations. To make the prediction feasible and efficient, a concentric motion distillation approach is developed by distilling the scene motion with key points. Finally, a Graph Convolutional Network is employed to predict the motions of key points, enabling the rendering of photorealistic images of future scenarios. Our framework shows outstanding performance on both synthetic and real-world datasets, demonstrating its efficacy in predicting and rendering future environments. Code is available on the project webpage:
  \urlstyle{tt}
  \textcolor{url_color}{\url{https://zju3dv.github.io/gaussian-prediction}}.
\end{abstract}



\begin{teaserfigure}
    \begin{center}    
    \includegraphics[width=0.85\linewidth, trim={0 0 0 0}, clip]{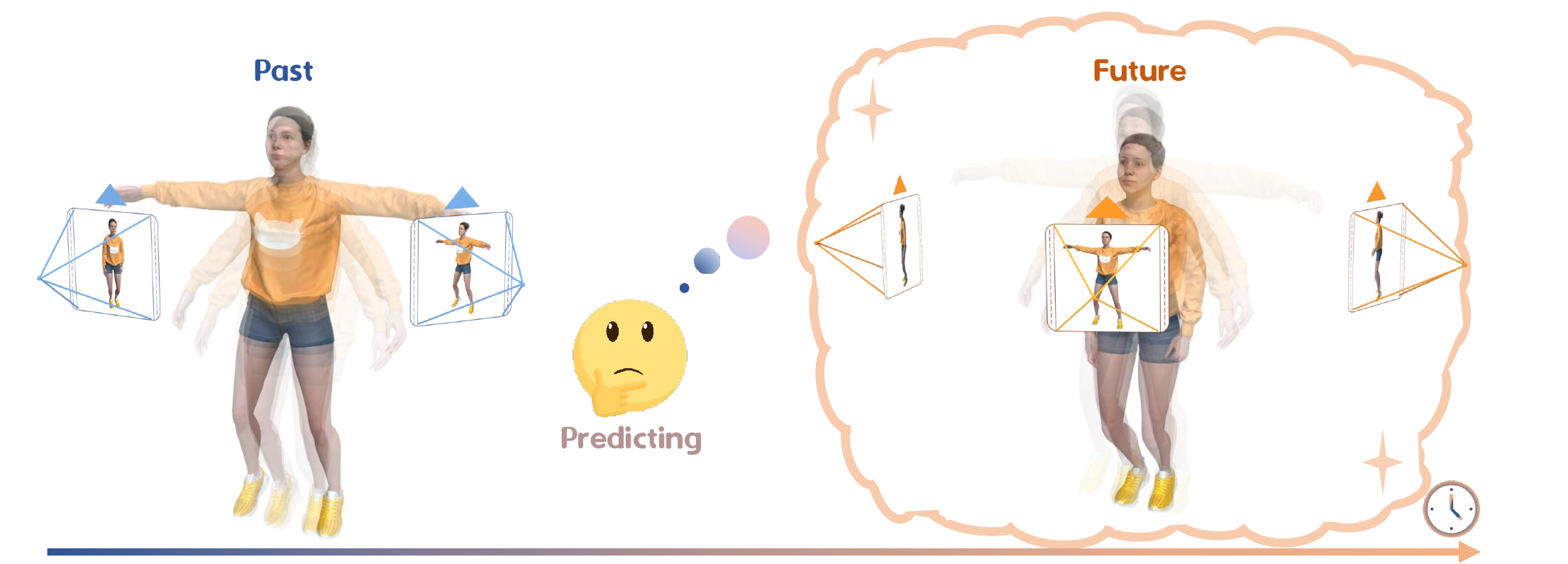}
    \end{center}
  \caption{GaussianPrediction can reconstruct the entire dynamic scene with high quality from monocular dynamic images and reasonably predict what will happen in the future. In addition, unlike 2D video prediction, our approach can synthesize novel view images at future moments.}
  \label{fig:teaser}
\end{teaserfigure}


\maketitle

\section{Introduction}
\label{sec:Introduction}
% 1. Introduce the background and explain the new task
The ability to envision what is about to happen in the near future is critical for us, human beings, to survive in ubiquitous dynamic scenes, and equally crucial for computers to fulfill intelligent decision-making and navigation in complex 3D worlds. 
Specifically, such ability cannot be well achieved without: 1) predicting the dense motion in a short future time span, and 2) visualizing the scene in the future from arbitrary viewpoints.
Despite extensive efforts in computer vision and robotics, imparting similar skills to intelligent agents remains a significant challenge.

% 2. Explain most relevant current works including video prediction and novel view synthesis and their drawbacks
One family of approaches aligned with this objective is video prediction, which endeavors to forecast the future dynamics of a scene from a specific viewpoint, based on past observations from that same viewpoint \cite{oprea2020review,neimark2021video,kwon2019predicting}. Despite its potential, video prediction falls short in visualizing arbitrary viewpoints, thus constraining its effectiveness in understanding the near future. 
On the other hand, novel-view synthesis focuses on rendering images of a scene from a flexibly chosen viewpoint \cite{mildenhall2021nerf, zhang2022differentiable, 3DGaussian}, %however, 
but it does not incorporate the element of temporal prediction necessary for forecasting future states of the environment.
Approaching in native 3D, 3D point cloud prediction aims to extrapolate future 3D point clouds from a sequence of past scans \cite{mersch2022self, wang2023semantic}. This technique is particularly beneficial for decision-making in the context of intelligent vehicles; however, it lacks the capacity to generate high-quality images.

\begin{figure*}[!t]
    \centering
    % \vspace{-3mm}
        %jumpingjacks
    \begin{subfigure}{0.19\linewidth}
        \centering
        \caption{TiNeuVox}
        \includegraphics[width=0.99\linewidth]{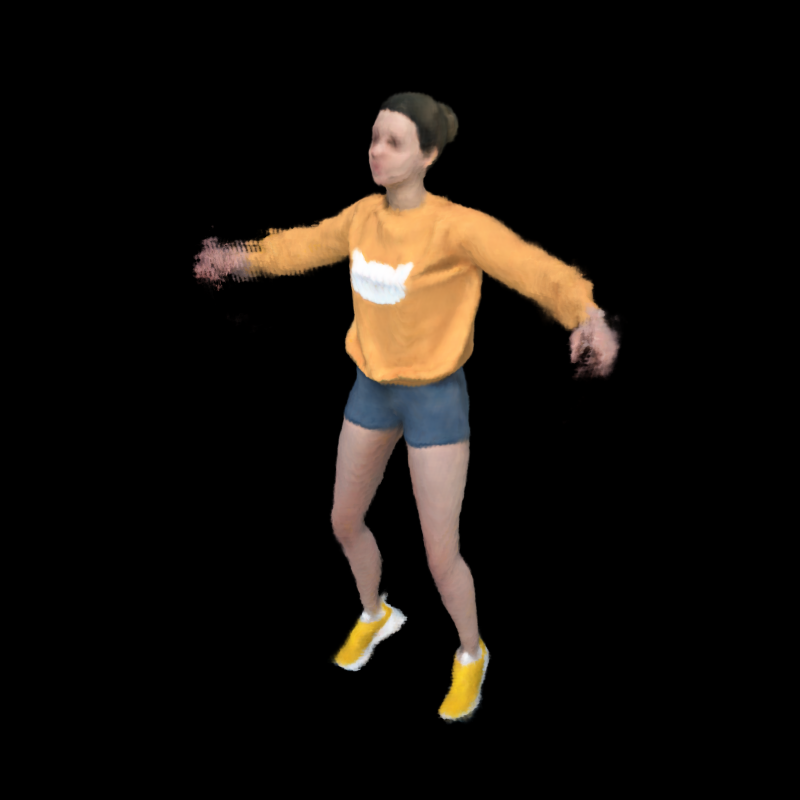}
    \end{subfigure}
    \begin{subfigure}{0.19\linewidth}
        \centering
        \caption{4D-GS}
        \includegraphics[width=0.99\linewidth]{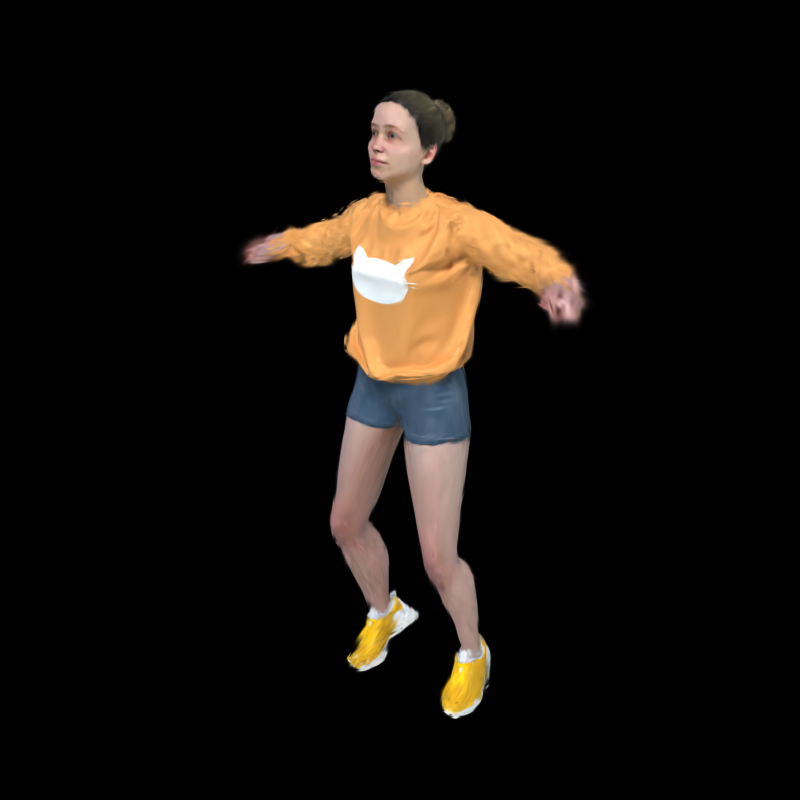}
    \end{subfigure}
    \begin{subfigure}{0.19\linewidth}
        \centering
        \caption{Deform-GS}
        \includegraphics[width=0.99\linewidth]{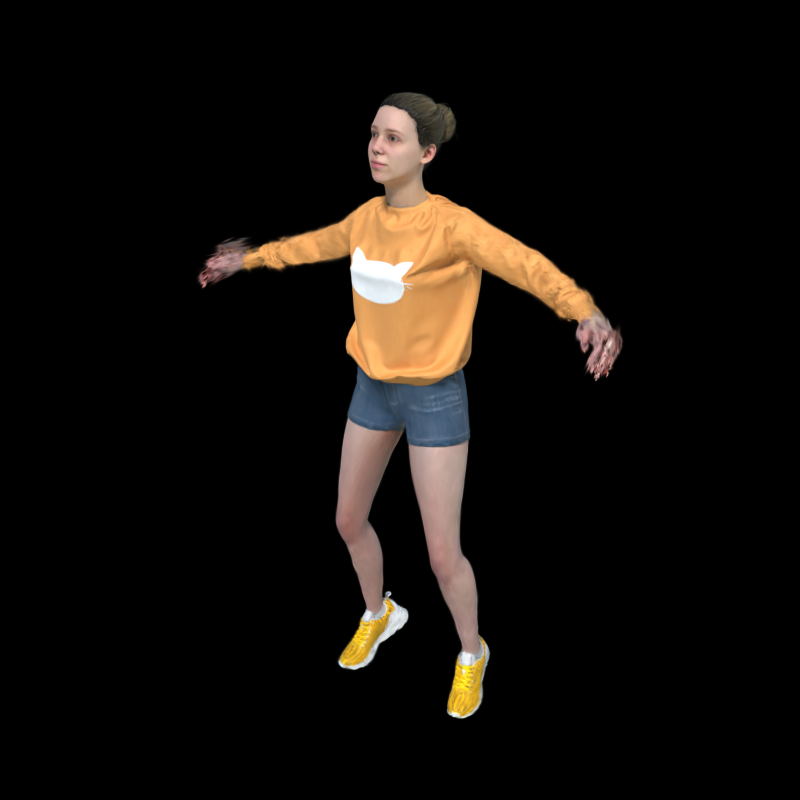}
    \end{subfigure}
    \begin{subfigure}{0.19\linewidth}
        \centering
        \caption{Ours}
        \includegraphics[width=0.99\linewidth]{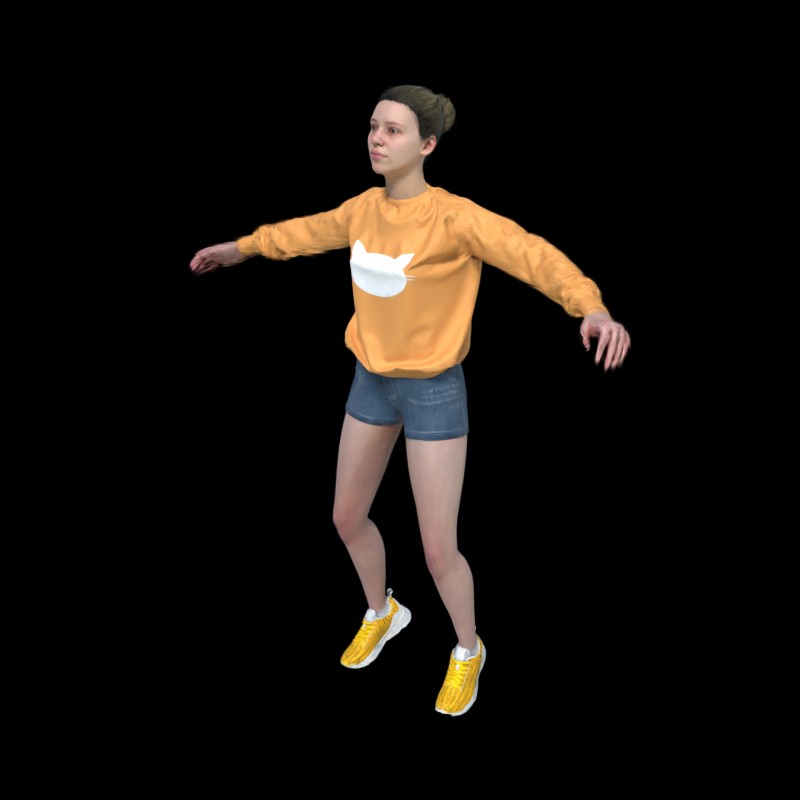}
    \end{subfigure}
    \begin{subfigure}{0.19\linewidth}
        \centering
        \caption{Ground Truth}
        \includegraphics[width=0.99\linewidth]{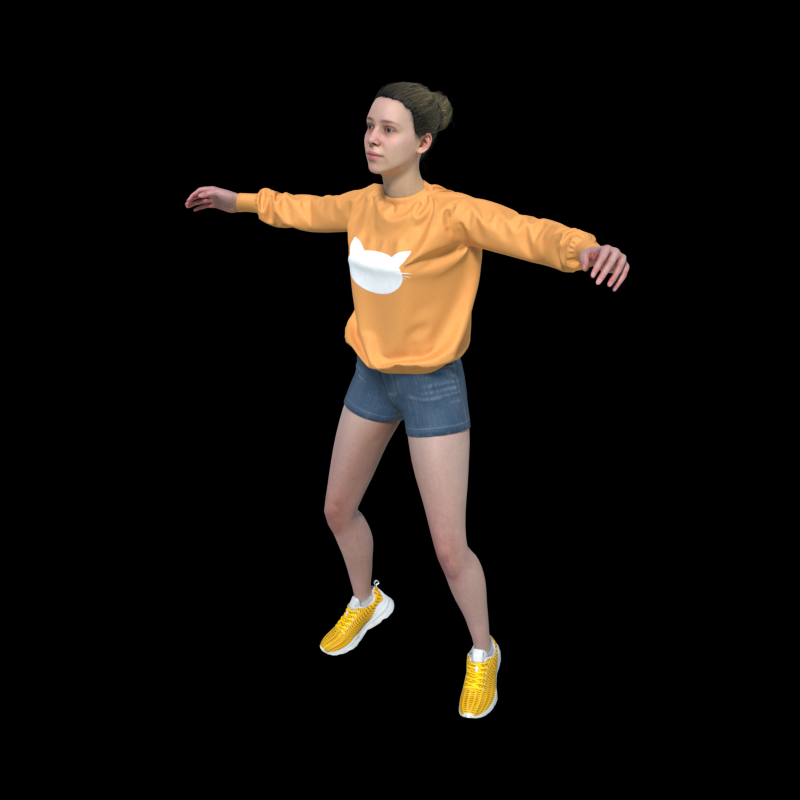}
    \end{subfigure}

        %standup
    \begin{subfigure}{0.19\linewidth}
        \centering
        \includegraphics[width=0.99\linewidth]{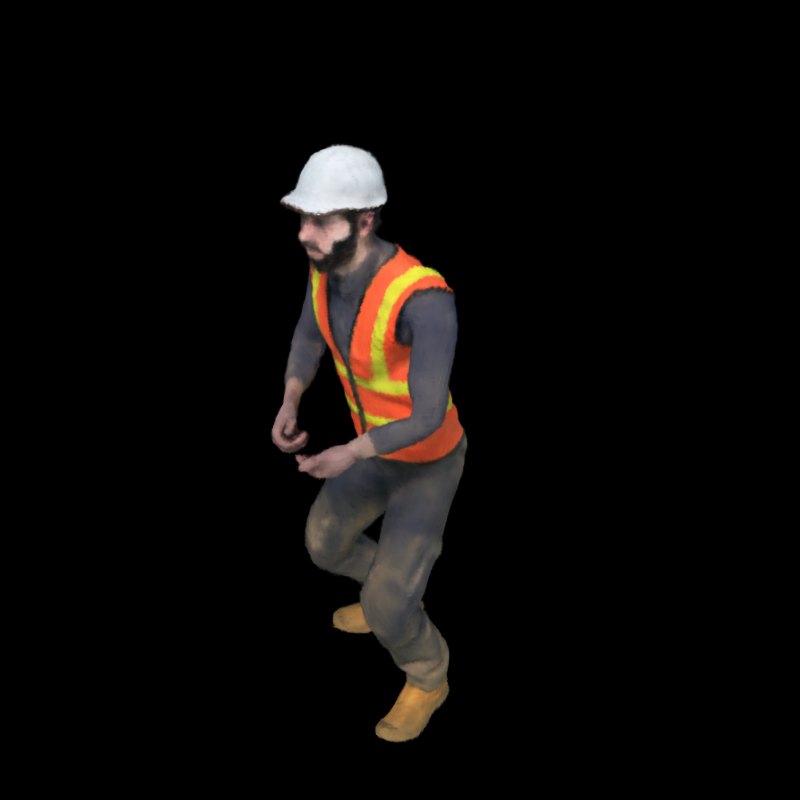}
    \end{subfigure}
    \begin{subfigure}{0.19\linewidth}
        \centering
        \includegraphics[width=0.99\linewidth]{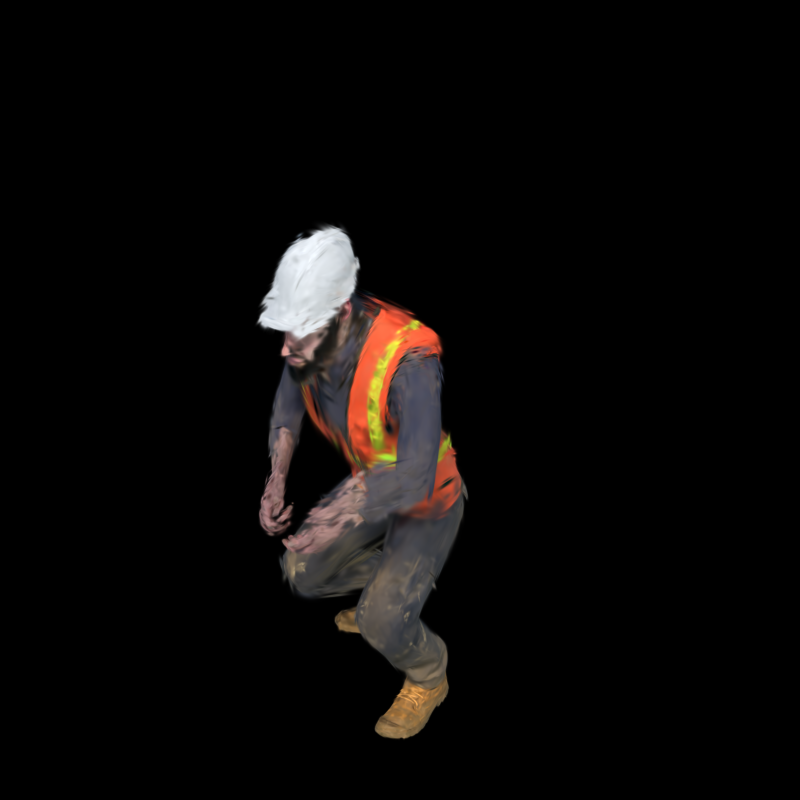}
    \end{subfigure}
    \begin{subfigure}{0.19\linewidth}
        \centering
        \includegraphics[width=0.99\linewidth]{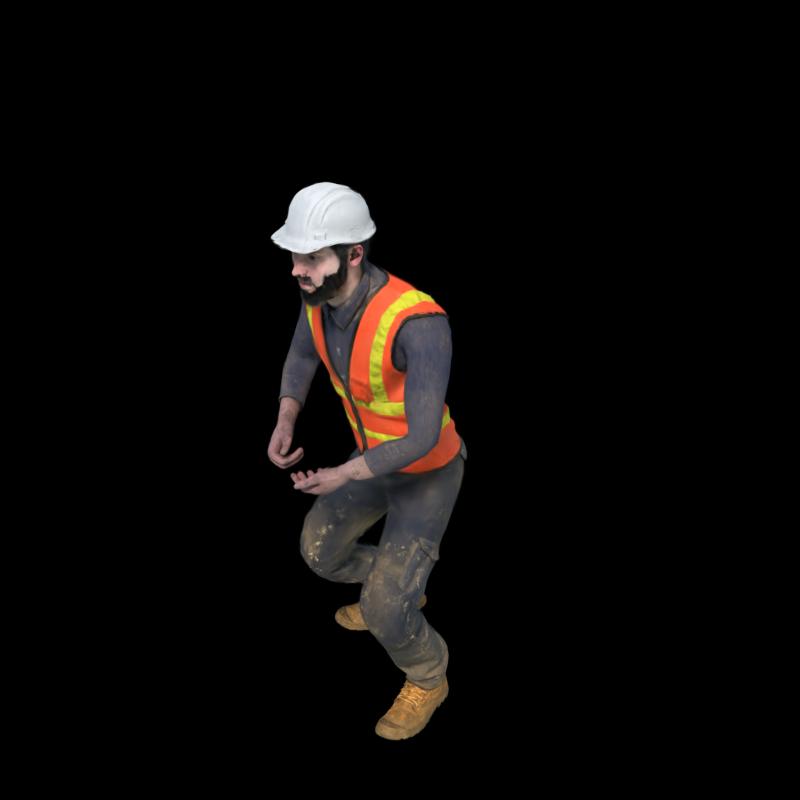}
    \end{subfigure}
    \begin{subfigure}{0.19\linewidth}
        \centering
        \includegraphics[width=0.99\linewidth]{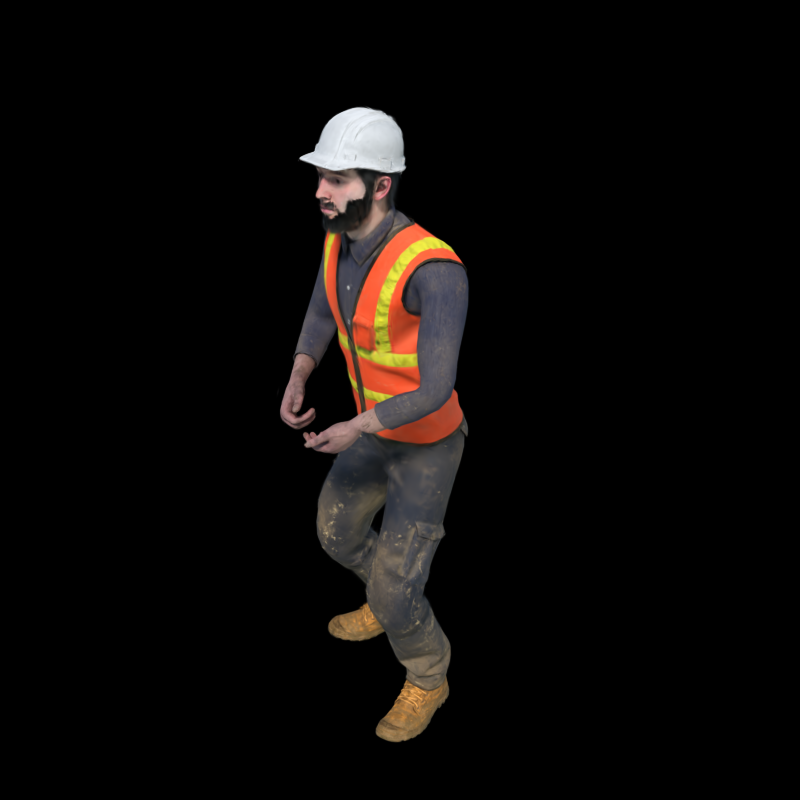}
    \end{subfigure}
    \begin{subfigure}{0.19\linewidth}
        \centering
        \includegraphics[width=0.99\linewidth]{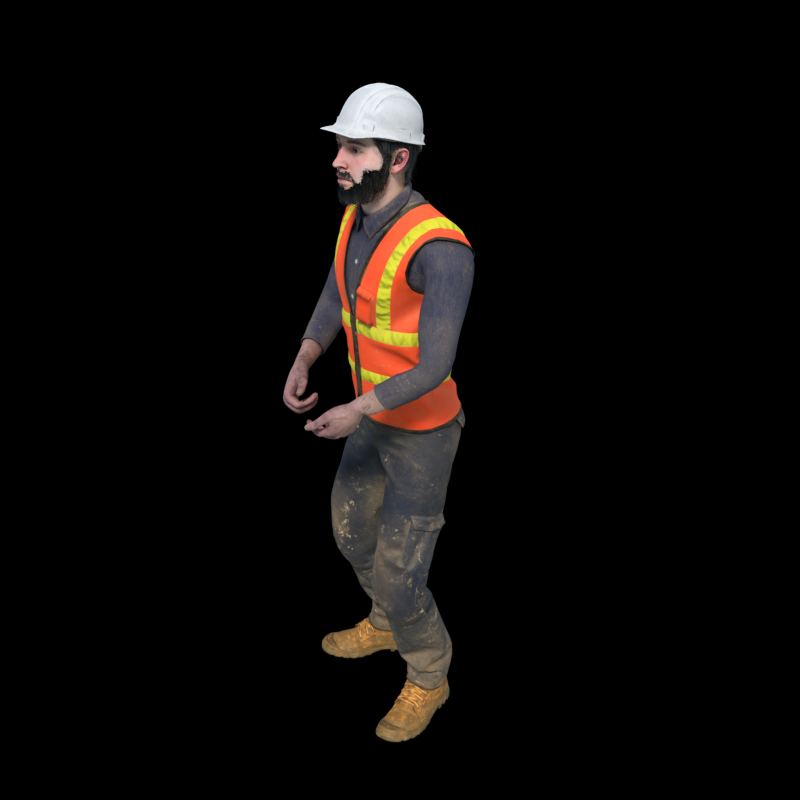}
    \end{subfigure}
    
    % \captionsetup{labelsep=period, labelfont=bf}
    % \vspace{-3mm}
    \caption{To distinguish from reconstructions results, we show prediction results in black background. Qualitative results on real-world scenes. We compare our methods with TiNeuVox~\cite{TiNeuVox}, 4D-GS~\cite{4D-Gaussians}, and Deform-GS~\cite{Deformable-Gaussian}. Our method not only renders highly detailed novel views but also predicts motion close to ground truth. Please refer to our supplementary video for more comparisons.}
    \label{fig:D-nerf-quality-compare}
\end{figure*}

% 3. Explain what we did and what's the challenge
In this paper, we introduce GaussianPrediction, a novel framework that builds on 3D Gaussian representations for dynamic scene modeling and future scenario synthesis in dynamic environments. Given video observations of dynamic scenes, GaussianPrediction is capable of forecasting potential future scenarios from any viewpoint. Although the 3D Gaussian representation \cite{3DGaussian} appears to inherently bridge novel-view synthesis and motion prediction, designing such a system is far from straightforward. Firstly, temporal modeling is needed for the 3D Gaussian representation to address both general motions and irreversible deformations in dynamic scenes, which is absent in concurrent works involving dynamic 3D Gaussian modeling \cite{4D-Gaussians,  Deformable-Gaussian}. Furthermore, accurately representing a scene requires a substantial number of 3D Gaussians, and their motions must be predicted in a cohesive manner to facilitate effective future projection and image rendering. However, prediction inaccuracies in even a small subset of these Gaussians can significantly degrade the quality of the rendered images (as shown in Fig.~\ref{fig:D-nerf-quality-compare}). 

% 4. Explain how we solve the problem
Drawing inspiration from D-NeRF ~\cite{D-NeRF} and its variants~\cite{Robust-dynamic-NeRF}, we first build the 3D Gaussian canonical space with deformation modeling to capture the appearance and geometry of the dynamic scene. We also design additional lifecycle properties for the Gaussians to model the irreversible deformations in a dynamic scene, such as those occurring on broken surfaces.
To efficiently control the motion of the entire scene's Gaussians, we employ a novel concentric motion distillation approach that utilizes key points to distill scene motions from the learned deformation fields. This markedly reduces the complexity of the following future prediction model, cutting down the number of nodes requiring prediction from hundreds of thousands to just a few hundred. Such a strategy makes the forecasting of future scenarios more efficient and feasible.
In enhancing the model's capability to accurately represent both appearances and motions with prediction capability, we select key points through feature clustering in a hyper-canonical space. This space is uniquely designed to encode both spatial and motion distances, thereby mitigating artifacts that often arise from discontinuous motion fields along object boundaries.
Finally, our framework leverages a Graph Convolutional Network (GCN) to predict the future motion of these 3D key points. This prediction, in turn, drives the anticipated deformation of the entire scene, enabling our system to forecast dynamic environments and synthesize future photorealistic images freely. 

% 5. Summary
Our contributions can be summarized as follows:
\begin{itemize}
    \item We present GaussianPrediction, a novel framework that innovatively integrates 3D Gaussian representations with dynamic scene modeling and future scenario synthesis, leveraging video observations to forecast scenes shortly from any viewpoint.
    \item We develop a novel canonical space of 3D Gaussians with lifecycle properties, which can model both general motions and irreversible deformations in dynamic scenes, offering a more comprehensive representation of temporal dynamics.
    \item We introduce a novel future prediction strategy based on the concentric motion distillation, which enhances the efficiency and robustness.
    \item Experiments on both synthetic and real-world datasets demonstrate the efficacy of our proposed framework in predicting and rendering future scenarios.
\end{itemize}

\section{Related Work}
\label{sec:Related}
\subsection{4D Novel View Synthesis}
Several studies address the reconstruction of dynamic scenes and the generation of free viewpoint renderings, employing explicit mesh representations~\cite{newcombe2015dynamicfusion, dou2016fusion4d, orts2016holoportation, broxton2020immersive}, depth estimations ~\cite{bansal20204d, yoon2020novel} or implicit neural volumes ~\cite{lombardi2019neural}. 
Harnessing its capability to deliver photorealistic novel renderings, Neural Radiance Field (NeRF)~\cite{mildenhall2021nerf} has been incorporated into 4D dynamic scene reconstructions~\cite{pumarola2021d, du2021neural, park2021nerfies, li2022neural, attal2023hyperreel}, spanning various tasks such as monocular video reconstructions~\cite{gao2021dynamic, tretschk2021non, li2021neural, li2023dynibar}, scene editings~\cite{Hyper-NeRF, kania2022conerf, Editable-NeRF}, human reconstructions~\cite{peng2021neural, peng2021animatable, weng2022humannerf, zielonka2023instant}, fast reconstructions and renderings~\cite{fang2022fast, fridovich2023k, shao2023tensor4d, cao2023hexplane, song2023nerfplayer, geng2023learning, lombardi2021mixture, peng2023representing}, as well as generalizable renderings~\cite{lin2022efficient, lin2023im4d}. Dynamic point clouds~\cite{zhang2022differentiable, zheng2023pointavatar, xu20234k4d} have garnered significant attention, particularly due to their rapid rendering speed. Notably, 3D Gaussian Splatting ~\cite{3DGaussian}  has emerged as a technique that combines swift reconstruction and rendering speeds, all while preserving exceptional rendering quality. 
Hence, a series of studies have been undertaken to broaden the applicability of 3D Gaussian Splatting to dynamic reconstruction scenarios. This has been achieved through explicit extensions of time-variant Gaussian features~\cite{luiten2023dynamic, yang2023real} or the utilization of implicit deformation fields ~\cite{4D-Gaussians, Deformable-Gaussian}. By combining explicit 3D Gaussian representations with implicit neural representations like MLPs~\cite{Deformable-Gaussian} or HexPlanes~\cite{4D-Gaussians}, these endeavors demonstrate exceptional qualities in novel view synthesis. They achieve interactive rendering frame rates and offer flexible editing capabilities, including object insertions.

\subsection{Dynamic Prediction}
Generating successive frames following given video sequences~\cite{oprea2020review} proves invaluable in intelligent decision-making. Approaches that combine 3D-CNNs, LSTMs, or Transformers ~\cite{wang2018eidetic, vondrick2016generating, villegas2018hierarchical, geng2023learning, girdhar2021anticipative, neimark2021video}seamlessly integrate spatial and temporal information during prediction. VAE-based methods~\cite{babaeizadeh2017stochastic, denton2018stochastic}, utilizing stochastic latent samples, yield diverse future predictions. Another line of work~\cite{lu2017flexible, kwon2019predicting}  employs GANs pre-trained on prior data to generate future predictions by analyzing historical frames. Other works~\cite{yang2023diffusion, hoppe2022diffusion} also produce realistic video predictions and infillings leveraging diffusion models.

Motion or dynamic predictions of 4D data inputs like human skeletons can also be conducted by RNNs~\cite{martinez2017human, corona2020context}, VAEs~\cite{petrovich2021action}, GANs~\cite{barsoum2018hp, martinez2017human} or diffusion models~\cite{alexanderson2023listen, barquero2023belfusion}. In the literature, there are also studies demonstrating predictions for scene-level dynamic point clouds, such as LiDAR scans ~\cite{mersch2022self} or common point clouds~\cite{wang2023semantic}. Approaches based on graph convolutional networks (GCN)~\cite{mao2019learning, sofianos2021space} exhibit excellent generalizing abilities while providing rapid prediction speeds. Nonetheless, these methods fall short in rendering photorealistic novel views compared to NeRFs' or 3D Gaussian Splattings' derivatives. In contrast, our GCN-based method excels in both forecasting reasonable scene dynamics based on historical frames and rendering high-quality novel views.

\section{Preliminaries}
\label{sec:Preliminary}
3D Gaussian Splatting~\cite{3DGaussian} employs a substantial number of explicit 3D Gaussians to represent a static 3D scene. Each 3D Gaussian $G$ is defined by a full covariance matrix $\Sigma$ and a center location $\mu$:
\begin{equation}
    G(x) = e^{-\frac{1}{2}(x-\mu)^T\Sigma^{-1}(x-\mu)}.
\end{equation}
For differentiable rendering optimization, 3D Gaussian splatting decomposes $\Sigma$ into scaling matrix $S$ and rotation matrix $R$: $\Sigma = RSS^{T}R^{T}$, 
where $S$ and $R$ are stored by a 3D vector $s$ and  a quaternion $q$ respectively.
To project these 3D Gaussians to 2D image, given a viewing transformation $W$, we obtain the 2D covariance matrix $\Sigma'$ and 2D center location $\mu'$: 
\begin{equation}
    \Sigma' = JW\Sigma W^{T}J^{T},
    \mu' = JW\mu,
\end{equation}
where J is the Jacobian of the affine approximation of the projective
transformation. Then we can use the neural point-based $\alpha$-blending to render the color $C$ of each pixel with $N$ ordered 3D Gaussians:
\begin{equation}
    C = \sum_{i \in N}T_ic_i\alpha_i,
\end{equation}
where $T_i, \alpha_i$ are calculated as:
\begin{equation}
    \begin{split}
    T_i &= {\textstyle \prod_{j=1}^{i-1}(1 - \alpha_j)}, \\
    \alpha_i &= \sigma_ie^{-\frac{1}{2}(x-\mu')^T\Sigma'^{-1}(x-\mu')}.
    \end{split}
\end{equation}
Here $\sigma_i$ is the opacity of the 3D Gaussian. Therefore, the 3D scene can be represented by the parameter set $P$ of 3D Gaussians, where $P = \{G_i: \mu_i, q_i, s_i, c_i, \sigma_i \}$.

\section{Method}
\label{sec:Method}
\begin{figure*}[t]
  \centering
  \includegraphics[width=0.95\linewidth, trim={0 0 0 0}, clip]{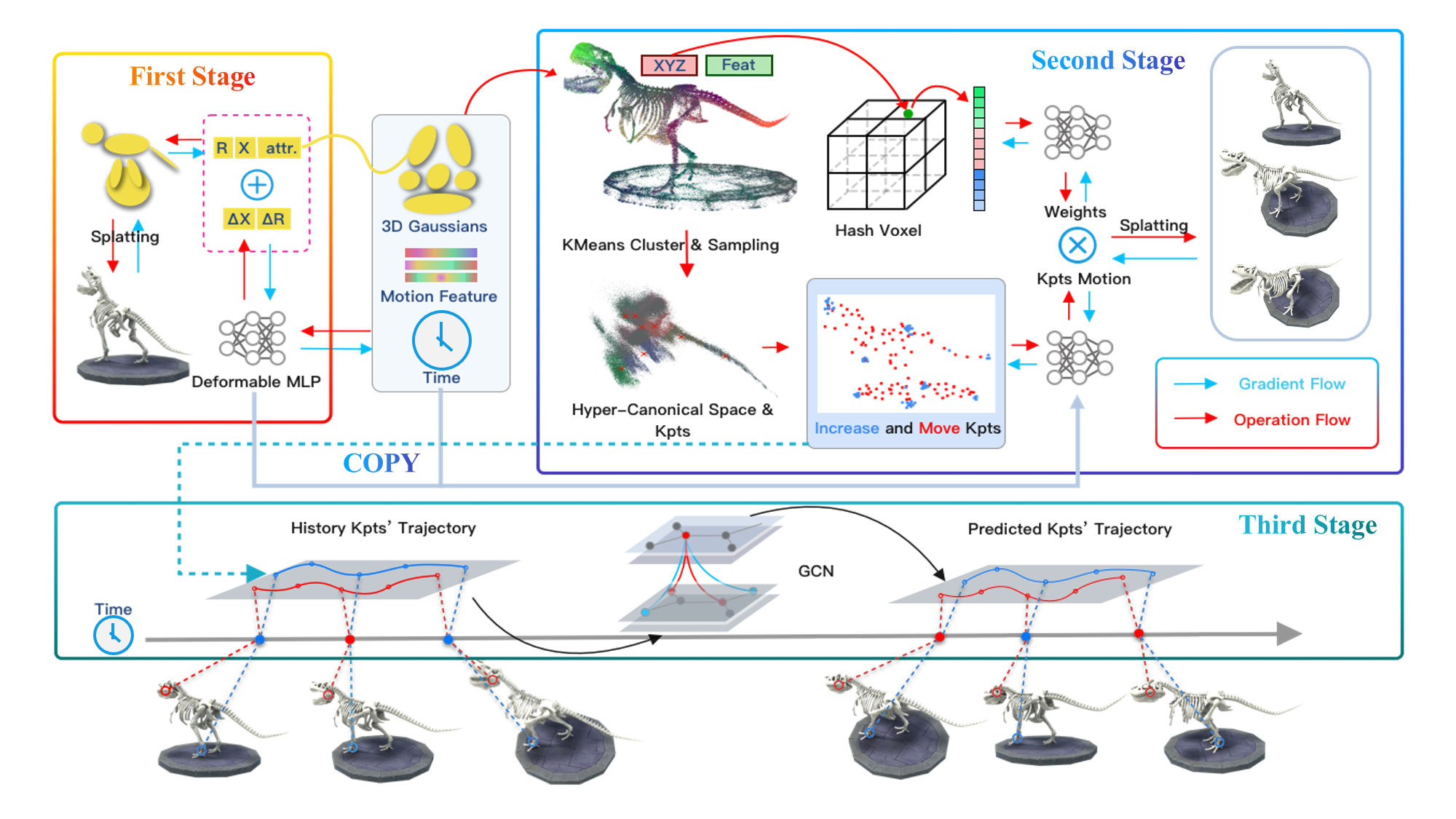}
  \caption{Optimization start with the initial 3D Gaussians. We then optimize the parameters of the 3D Gaussians, motion feature, and deformable MLP to build a Hyper-Canonical space. Next in the second stage, we first initialize the key points in the Hyper-Canonical space by a K-Means algorithm. Then we learn the time-independent weights for each Gaussian and deform the 3D Gaussian by key points motion. We employ a GCN (Graph Convolutional Network) to learn the relationships between key points, thereby predicting the future motion of key points, and rendering future scenes from a novel view.}
  \label{fig:pipeline}
\end{figure*}

Given a collection of images of a scene captured at different time instances from a monocular camera, GaussianPrediction aims to reconstruct the dynamic scenes and forecast future scenarios. %predict what will happen next. 
As shown in Fig.~\ref{fig:pipeline}, GaussianPrediction consists of three stages. At first, we recover the canonical space of 3D Gaussians with deformation fields to model the dynamic scenes from input images (Sec.~\ref{ssec:canonical}). To model irreversible deformations (\eg, cutting fruits or splitting cookies), we extend the 3D Gaussian with a novel lifecycle property. In the second stage, we employ a novel concentric motion distillation approach with key points to distill scene motion, significantly reducing the complexity of forecasting by decreasing the parameters from hundreds of thousands to a few hundred  (Sec~\ref{ssec:keypoints}). 
Finally, in the third stage, we adopt Graph Convolutional Networks (GCN) to predict the future motion of these 3D key points, effectively forecasting the entire scene's deformation (Sec.~\ref{ssec:gcn}). 

\subsection{Dynamic Modeling with Canonical Space}
\label{ssec:canonical}
\paragraph{Hyper-canonical space of 3D Gaussians.} 
One widely used strategy for reconstructing dynamic scenes is to build a canonical space and subsequently deform all the 3D information within this canonical space to match different time instances. Previous methods~\cite{D-NeRF, Deformable-Gaussian, Robust-dynamic-NeRF} employed Multilayer Perceptrons (MLP) to encode temporal deformation, utilizing the 3D location $x$ and time $t$ as inputs. However, due to the inherent similarity in spatial characteristics among adjacent 3D positions, these approaches may result in blurring when handling different motion patterns in neighboring locations. To address this problem, we utilize a motion feature $m$ with dimension $d$ for each Gaussian to encode the motion information. Given any timestamp $t$, we can obtain the deformation of each 3D Gaussian's center location $\mu$ and rotation $q$ from the canonical space to the moment $t$:
\begin{equation}
\label{eq:deformable}
    \Delta \mu^t, \Delta q^t = D(\gamma(\mu), m, \gamma(t)),
\end{equation}
where $D$ is a Multilayer Perceptron (MLP) and $\gamma$ denotes the positional encoding with frequency $L$:
\begin{equation}
    \gamma(x) = (sin(2^l\pi x), cos(2^l\pi x))^{L-1}_{l=0}.
\end{equation}
Therefore our deformed 3D Gaussians $P_t$ at timestamp $t$ can be defined as: $P^t = \{G_i^t: (\mu_i + \Delta \mu^t_i), (q_i \otimes \Delta q_i^t ), s_i, c_i, \sigma_i, m_i \}$, where $\otimes$ represents quaternion multiplication. 
Then we define the hyper-canonical space $C_h$ as the following:
\begin{equation}
    C_h = \{ (\mu, m) | \mu \in \mathbb{R}^3, m \in \mathbb{R}^d \}.
\end{equation}
However, we found that optimizing both $\mu$ and the deformable MLP can easily get stuck in local optima. Therefore, we propose to introduce an annealing noise $\varepsilon$ on $\mu$:
\begin{equation}
    \varepsilon(i) = \mathcal{N}(0, 1) \cdot N_s \cdot (1 - min(1, \frac{i}{10000})),
\end{equation}
where $i$ denotes the current training iteration and $N_s$ is the scaling factor. In our experiments, we observed that decaying noise effectively guides the optimization of $\mu$ away from local optima. This leads to a more uniform distribution of $\mu$ in space and enhances the rendering quality of our model (See Sec.~\ref{ssec:ablation}).

\paragraph{Lifecycle of Gaussians.} 
Temporal motion often triggers situations where a portion of the original surface disappears (\eg, gluing up toy pieces together) or new surfaces emerge (\eg, cutting a lemon into two halves).
This deformation is irreversible, which is different from the general deformation in that correspondences exist over the whole sequence. An intuitive idea is that we should allow 3D Gaussians to exhibit the same property—being renderable until a certain point in time, after which they lose their rendering abilities. Therefore, we propose to add a lifecycle $\psi$ to the opacity of 3D Gaussians:
\begin{equation}
% \vspace{-0.5em}
\begin{split}
    \psi(G_i, t) &= \frac{1}{1 + e^{(-10\Delta_o (G_i, t))}}, \\
    \Delta_o &= D_o(\gamma(\mu_i), m_i, \gamma(t)).
\end{split}
\end{equation}
Here $\Delta_o$ is calculated by the opacity deformable MLP $D_o$. We then multiply this lifecycle $\psi$ with the opacity of the 3D Gaussian which makes the opacity mostly either 0 or 1 in the majority of cases, indicating that Gaussians are involved in rendering at certain moments while being invisible at other times. Our experiment shows that this strategy efficiently improves the rendering quality both quantitatively and qualitatively (See Sec.~\ref{ssec:ablation}). %(Sec.~\ref{sssec:ablation_lifecycle}).

%\subsection{Driven by Learnable Key Points}
\subsection{Concentric Motion Distillation with Key Points}
\label{ssec:keypoints}
A direct way to predict the future deformation of the scenes is to extrapolate the input time 
$t$ to $D$ in Eq.~\ref{eq:deformable}. However, each 3D Gaussian in the canonical space is independent and unconstrained, and as a result, direct extrapolation would cause 3D Gaussians to lose their original geometric properties. To address this problem, we design a key points driven framework inspired by~\cite{Editable-NeRF} to deform the whole 3D Gaussians by key points $K = \{k_i = (\mu_i^k \in \mathbb{R}^3, m_i^k \in \mathbb{R}^d )\}, i \in \{1,2,..., N_k\}$ defined in the hyper-canonical space. Here $\mu_i^k$ denotes the 3D position and $m_i^k$ is the motion feature of key points $k_i$. For each key point $k_i$, we calculate the 3DoF translation vector $T_i^t$ and 3DoF rotation quaternion $Q_i^t$ at time $t$ by the deformable MLP $D$ trained in Sec.~\ref{ssec:canonical}:
\begin{equation}
    T_i^t, Q_i^t = D(\gamma(\mu_i^k), m_i^k, \gamma(t)).
\end{equation}
Note that the representation of motion for key points here is identical to the representation of motion for 3D Gaussians in the first stage. Therefore, we can utilize the deformable MLP trained in the first stage as the deformable MLP for the second stage. Then we can render the image at time $t$ using the deformed 3D Gaussians $P_t^{\text{key}}$ which can be represented as:
\begin{equation}
\label{Eq: weights}
    \begin{split}
        P_t^{\text{key}} &= {\{G_i^t: (\mu_i + \Delta \mu^t_i), (q_i \otimes \Delta q_i^t ), s_i, c_i, \sigma_i, m_i \}},\\
        \Delta \mu^t_i &= \sum_{k\in K} (w^T_{i\leftarrow k} \cdot  T_{k}^t), \Delta q_i^t = \sum_{k\in K} (w^Q_{i\leftarrow k} \cdot  Q_{k}^t),
    \end{split}
\end{equation}
where $w^T_{i\leftarrow k}$, $w^Q_{i\leftarrow k}$ represent the translation and rotation weights of 3D Gaussian $G_i$ with respect to key point $k$.
In summary, our concentric motion distillation comprises three key steps:
\begin{enumerate}
\item \textbf{Initializing key points.} This foundational step will initialize $k_{init}$ key points in the hyper-canonical space.
\item \textbf{Adaptive increasing key points.} For those complex motion areas, the initial key points may not be enough. Therefore we need to increase the key points in complex motion areas adaptively. 
\item \textbf{Time-independent weights learning.} This step involves understanding how each key point affects each 3D Gaussian, which transforms the motion of key points into the motion of 3D Gaussians.
\end{enumerate}
In this way, we can deform more than 200k 3D Gaussians using hundreds of key points, which simplifies the prediction process.
Next, we will introduce the details of each step.

\paragraph{Initializing key points.} 
Once we have trained the hyper-canonical space, we can sample $k_{init}$ 3D points as the initial key points. Gaussians driven by the same key point should demonstrate both motion similarity and spatial proximity. To achieve this, we employ clustering techniques in the hyper-canonical space to organize 3D Gaussians into $k_{init}$ classes as $S = \{S_1, S_2, ..., S_{k_{init}}\}$. Formally, the objective is to find:
\begin{equation}
    \underset{S}{\text{argmin}} \sum_{i=1}^{K_{init}} \frac{1}{|S_i|} \sum_{x,y \in S_i}\left \| x - y \right \|^2,
\end{equation}
where x,y is the vector $\in \mathbb{R}^{3+d}$ in $C_h$. We then utilize the 3D center of each class as the initial position for the $k_{init}$ key points.

\paragraph{Adaptive increasing key points.} Following~\cite{3DGaussian}, we also present an adaptive increasing strategy to add key points near the complex motion 3D Gaussians. To identify the areas that require additional key points, we calculate the Gaussian gradient norm and select Gaussians with a norm greater than the gradient threshold. A large gradient means these Gaussians show poor reconstruction results. Therefore, we use FPS~\cite{FPS} uniformly downsampling the large gradient Gaussians by a factor of 100 to generate the locations for the newly added key points. However, to facilitate predicting scene motion, we limit the number of added points to avoid increasing complexity. Thus the maximum number of adaptive increasing key points is $N - k_{init}$.
In this way, our model can flexibly handle complex motions.

\paragraph{Time-independent weights learning.} 
In Eq.~\ref{Eq: weights}, we introduced the time-independent weights $w^T, w^Q$ and discussed how to drive the 3D Gaussian by key points motion. In this step, we will discuss how to learn these weights for each 3D Gaussian.

Given a 3D Gaussian $G_i$ in the canonical space, its motion is primarily influenced by the movements of the nearest $N_\text{near}$ key points, rather than those that are far away. As a result, we can represent the weights $w^T_i, w^Q_i$ for the Gaussian $G_i$ \wrt the key points as:
\begin{equation}
    w^T_i = \text{softmax}(\sum_{k \in K_{\text{near}}^i}^{} w^T_{i\leftarrow k}),
    w^Q_i = \text{softmax}(\sum_{k \in K_{\text{near}}^i}^{} w^Q_{i\leftarrow k}),
\end{equation}
where $k^i_\text{near}$ represents the $N_\text{near}$ key points closest to the $G_i$. However, we found that directly finding the nearest neighbors in space without considering motion information may not completely separate Gaussians with different motions but are spatially adjacent. To address this issue, we propose searching for the nearest key points to each Gaussian point in the Hyper-Canonical space $C_h$ taking into account both spatial proximity and motion similarity. Our experiments demonstrate the effectiveness of this nearest key points strategy which significantly reduces artifacts in the real-world dataset (See Sec.~\ref{ssec:ablation}).

EditableNeRF~\cite{Editable-NeRF} takes canonical coordinate $x$ as input and outputs a weight vector $w$ by a large MLP. However, this method is inefficient when the number of queries $x$ increases. Inspired by~\cite{Instant-ngp}, we present a novel method that uses hash encoding to map the coordinate $x$ to trainable feature vectors and then decode to $w$ by a tiny MLP. In this way, our model remains efficient even with a substantial increase in input coordinates.

\subsection{GCN-based Motion Prediction}
\label{ssec:gcn}
After the optimization in the second stage, we obtain several key points that encapsulate the distilled motion information of the scene. All the scene motion details are implicitly encoded within these key points. Considering the relationships between these key points, we utilize the capabilities of Graph Convolution Network~(GCN) to model and predict the dynamic movement patterns of key points within a given scene. At different time steps, we use the 3D positions of key points as supervision. We employ a GCN to extract relational features between key points across multiple frames. A single-layer MLP is then utilized to decode these features and predict the positions of key points at the next time step.  Then, we calculate the motion of each 3D Gaussian using Eq.~\ref{Eq: weights}, obtaining the predicted 3D Gaussians. Our approach allows for continuous predictions using a sliding window, leveraging past time steps to generate new predictions. Please refer to our supp. material for more details.
\begin{figure}[t]
  \centering
  \includegraphics[width=0.90\linewidth]{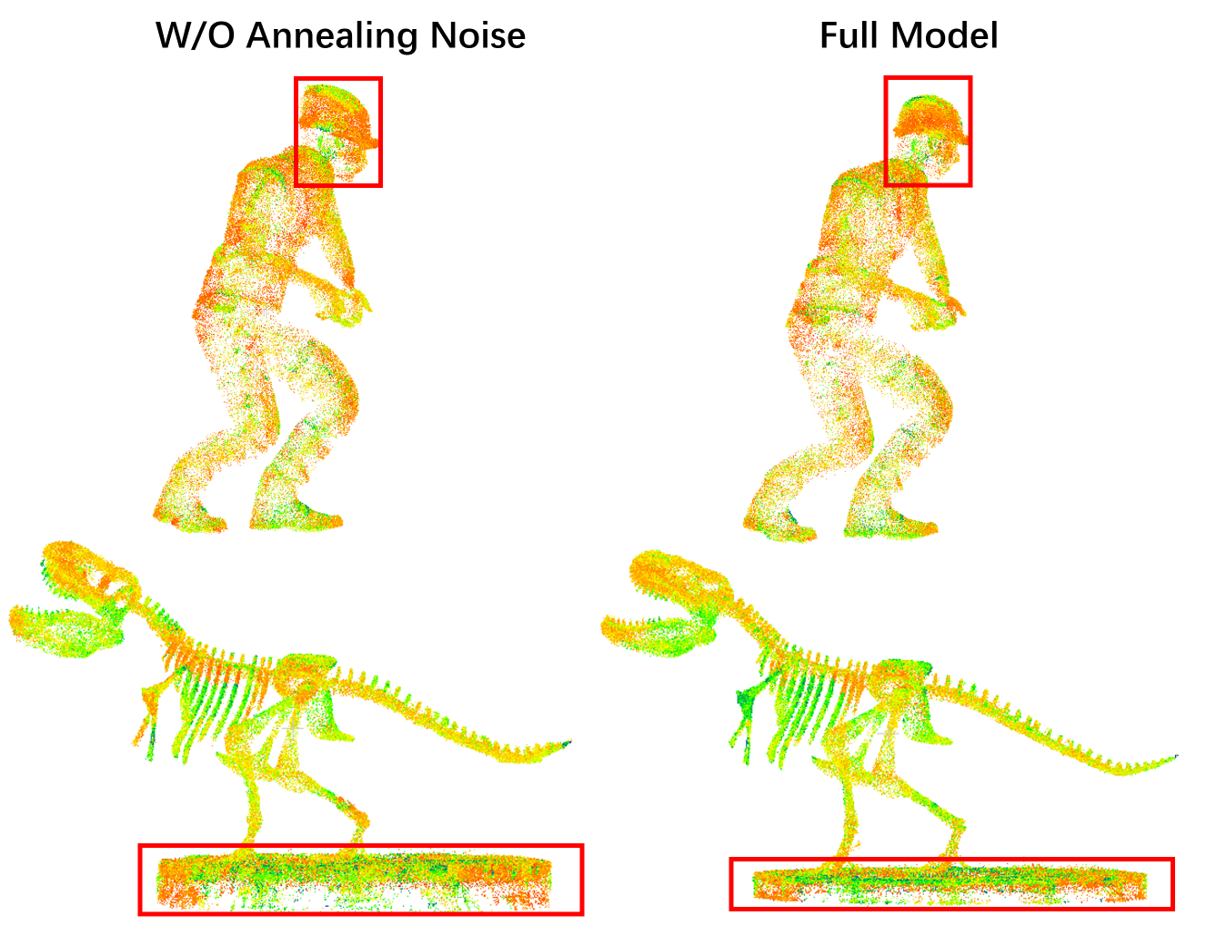}
  \caption{Canonical space point cloud with different training strategies.}
  \label{fig:ablation_pcd}
\end{figure}

\begin{figure}[!t]
    \centering
    \begin{subfigure}{0.48\linewidth}
        \centering
        \includegraphics[width=0.9\linewidth]{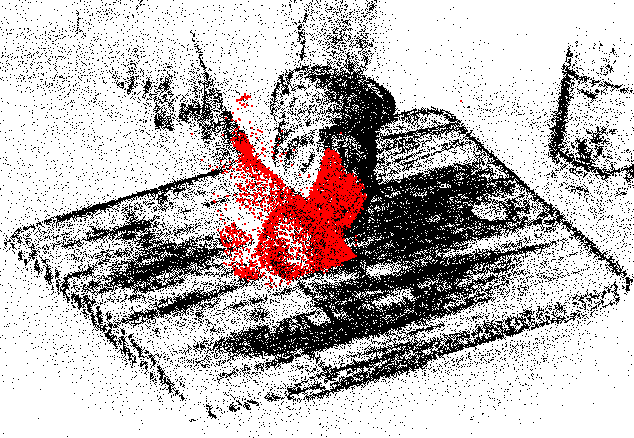}
        \caption{Searching nearest key points in 3D space.}
        \label{subfig:3D_select}
    \end{subfigure}
    \begin{subfigure}{0.48\linewidth}
        \centering
        \includegraphics[width=0.9\linewidth]{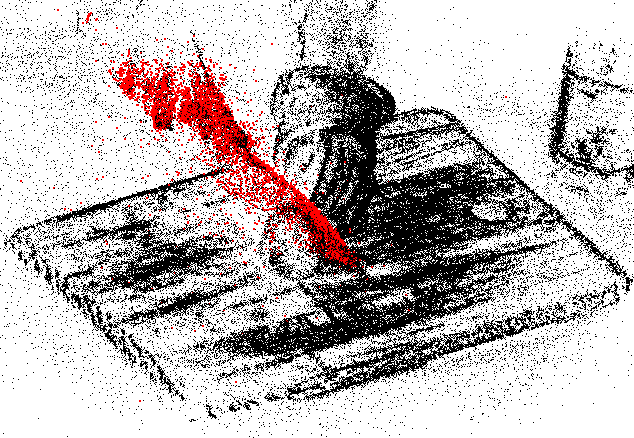}
        \caption{Searching nearest key points in Hyper-Canonical space.}
        \label{subfig:hyper_select}
    \end{subfigure}
    \caption{Influenced 3D Gaussians by a key point on the knife. We compare two different search methods and show influenced Gaussian points in \textcolor{red}{red}.}
    \label{fig:ablation_knn_pcd}
\end{figure}

\begin{figure*}[ht]
    \centering
    \begin{subfigure}{0.19\linewidth}
        \centering
        \includegraphics[width=0.99\linewidth]{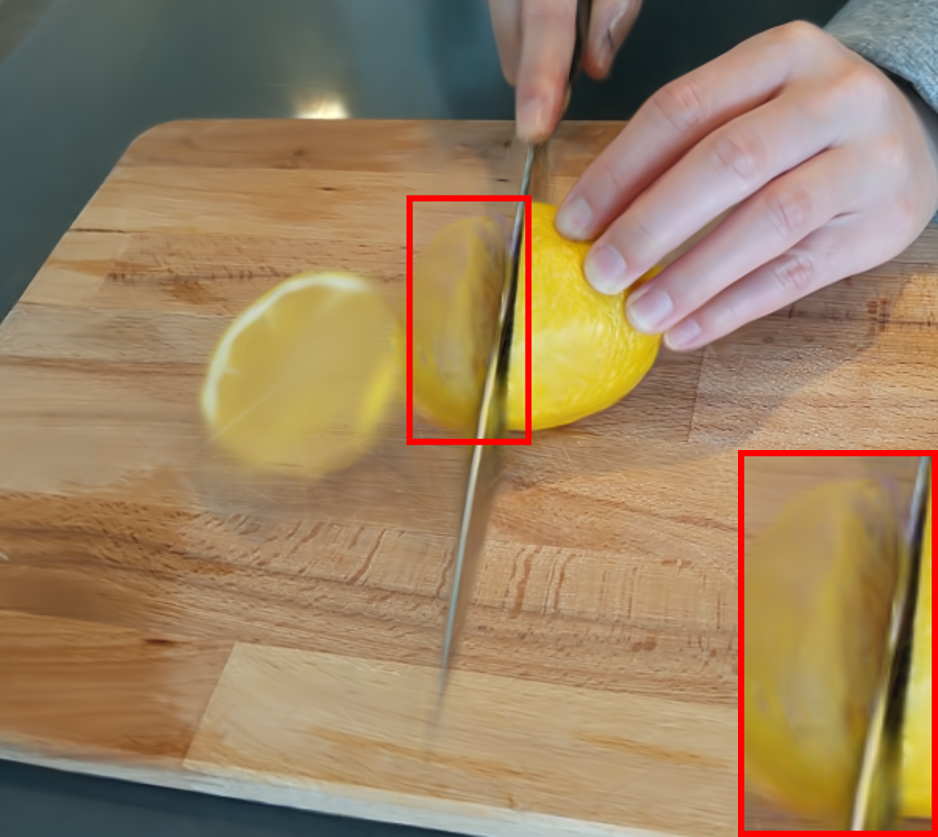}
        \caption{w/o Hyper Space Init.}
        \label{subfig:ablation_woInit}
    \end{subfigure}
    \begin{subfigure}{0.19\linewidth}
        \centering
        \includegraphics[width=0.99\linewidth]{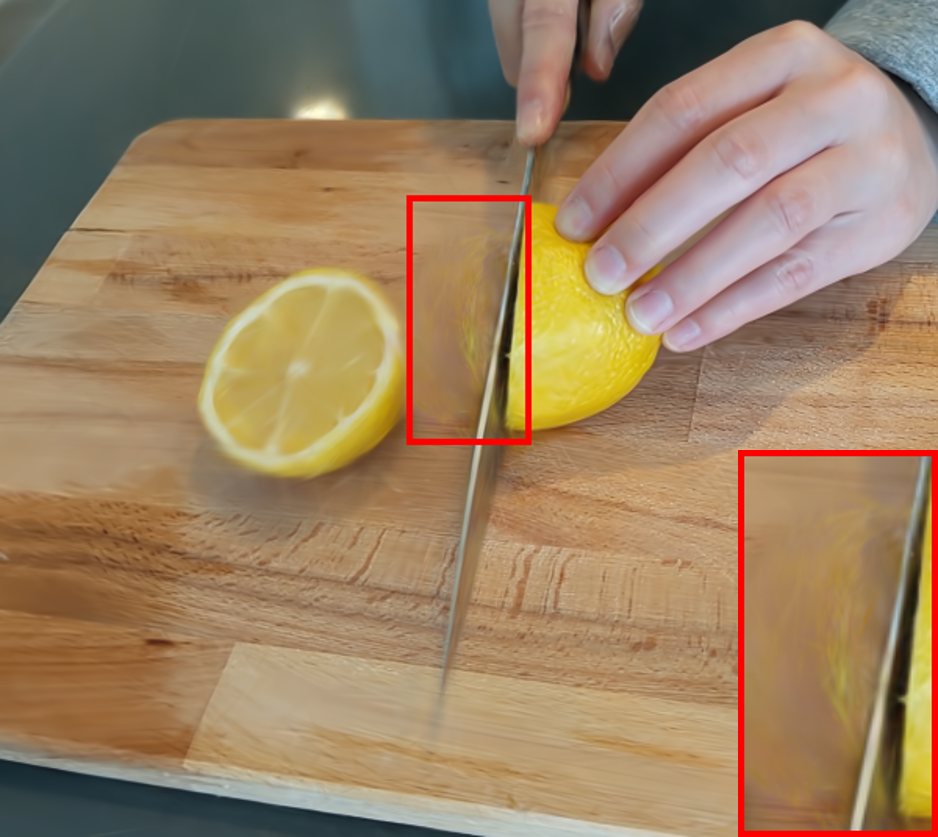}
        \caption{w/o Adap Increasing.}
        \label{subfig:ablation_woAdaptive}
    \end{subfigure}
    \begin{subfigure}{0.19\linewidth}
        \centering
        \includegraphics[width=0.99\linewidth]{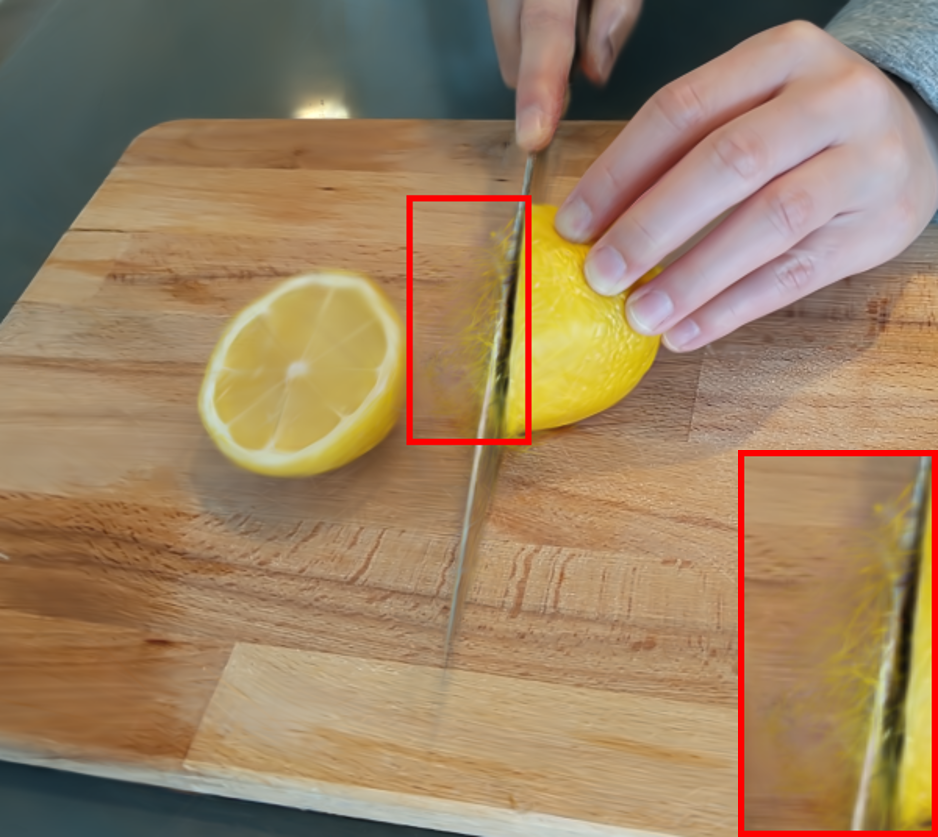}
        \caption{w/o Hyper Space K-NN.}
        \label{subfig:ablation_woClusterKnn}
    \end{subfigure}
    \begin{subfigure}{0.19\linewidth}
        \centering
        \includegraphics[width=0.99\linewidth]{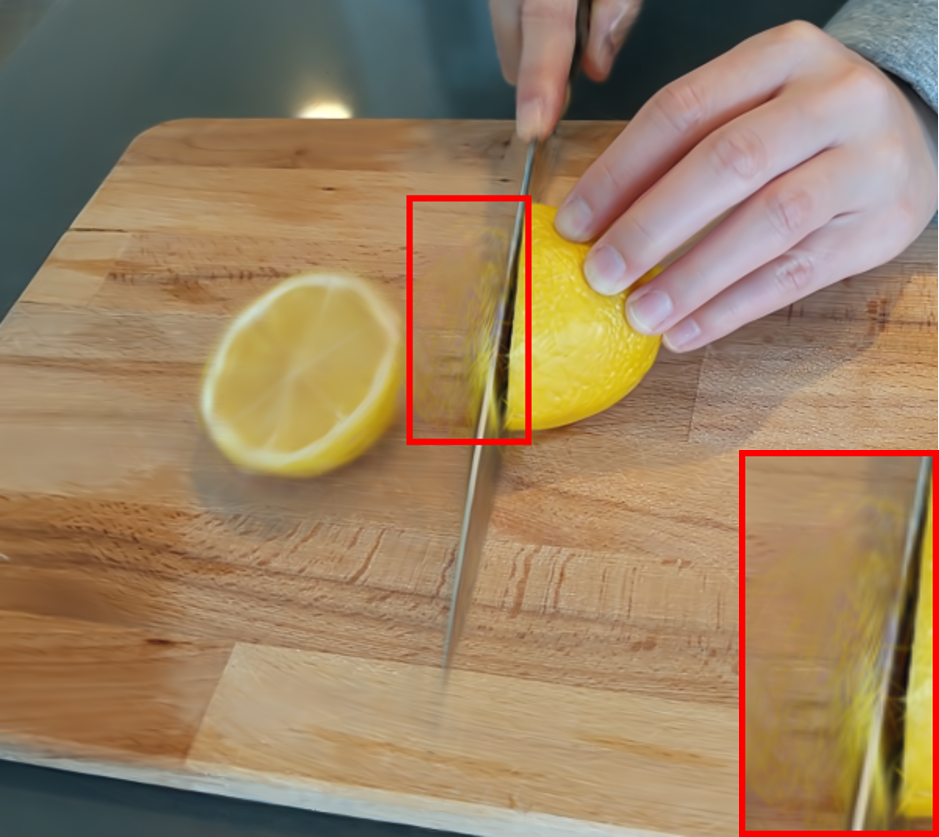}
        \caption{w/o Lifecycle.}
        \label{subfig:ablation_woLifecycle}
    \end{subfigure}
    \begin{subfigure}{0.19\linewidth}
        \centering
        \includegraphics[width=0.99\linewidth]{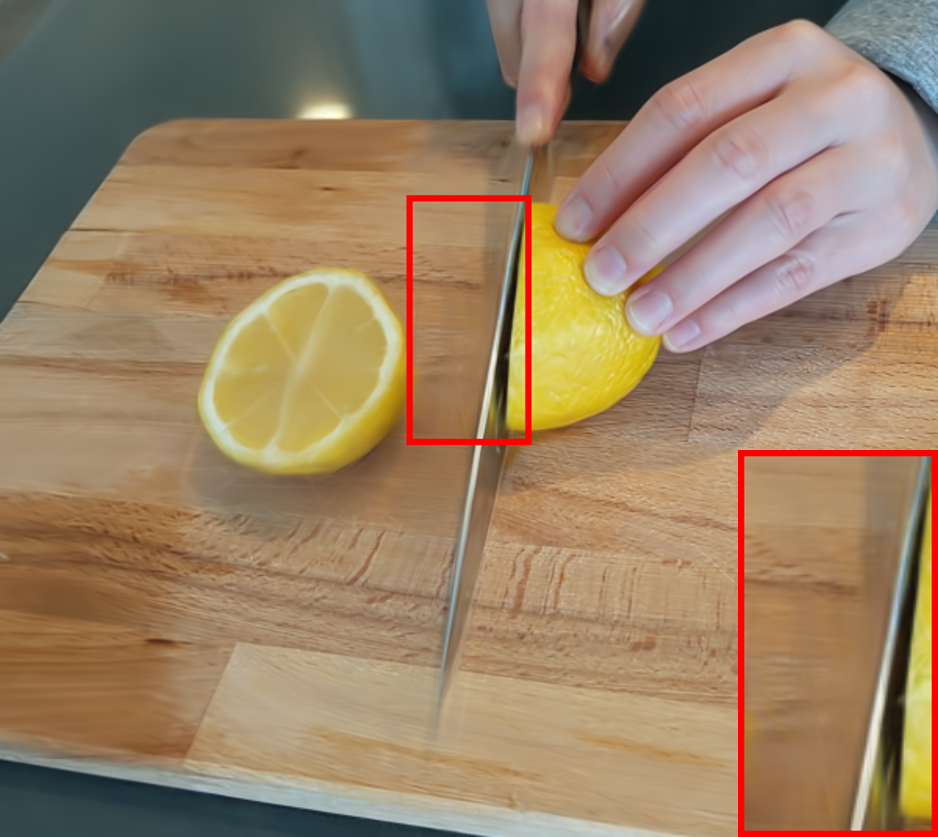}
        \caption{Full Model.}
        \label{subfig:ablation_full}
    \end{subfigure}
    \caption{We analyze the effectiveness of each component.}
    \label{fig:ablation}
\end{figure*}

\section{Experiments}
\label{sec:Exp}
\begin{table*}[t]
\centering
\caption{Quantitative motion prediction results comparison on D-NeRF dataset. Following HyperNeRF~\cite{Hyper-NeRF}, the average metrics are calculated using a weighted average. Best results are highlighted as \fs{\bf{first}},~\nd{\bf{second.}}}
% \vspace{-4mm}
\setlength{\extrarowheight}{2.5pt}
    \footnotesize
    \begin{tabular}{lcccccccccccc}
        \Xhline{1.2pt}
      & \multicolumn{3}{c}{\textbf{Trex}}        & \multicolumn{3}{c}{\textbf{Jumpingjacks}} & \multicolumn{3}{c}{\textbf{Bouncingballs}}    & \multicolumn{3}{c}{\textbf{Hellwarrior}} \\ \cmidrule(lr){2-4} \cmidrule(lr){5-7} \cmidrule(lr){8-10} \cmidrule(lr){11-13} 
        \multirow{-2}{*}{Method}  & PSNR($\uparrow$)     & SSIM($\uparrow$)   & LPIPS($\downarrow$)   & PSNR($\uparrow$)    & SSIM($\uparrow$)    & LPIPS($\downarrow$)    & PSNR($\uparrow$)   & SSIM($\uparrow$)  & LPIPS($\downarrow$)  & PSNR($\uparrow$)     & SSIM($\uparrow$)    & LPIPS($\downarrow$)    \\ \toprule
        TiNeuVox-B    & 20.72 & .9284 & .0751 & 19.87 & .9115
        & .0954 & 25.92 & .9677 & .0853 & 29.36 & .9097 & .1138 \\
        4D-GS         & 20.72 & .9401 & .0579 & 20.28 & .9176 & .0825 & \cellcolor[HTML]{FAC791}29.42 & .9753 & .0433 & \cellcolor[HTML]{F59194}31.48 & \cellcolor[HTML]{FAC791}.9266 & .0929 \\
        Deformable-GS & 20.81 & \cellcolor[HTML]{FAC791}.9426 & .0461 & 20.21 & .9150 & .0800 & 28.90 & \cellcolor[HTML]{FAC791}.9784 & \cellcolor[HTML]{FAC791}.0271 & 29.82 & .9141 & .0834 \\ \hline
        Ours-MLP      & \cellcolor[HTML]{F59194}21.51 & \cellcolor[HTML]{F59194}.9444 & \cellcolor[HTML]{F59194}.0452 & \cellcolor[HTML]{F59194}20.68 & \cellcolor[HTML]{F59194}.9194 & \cellcolor[HTML]{F59194}.0742 & \cellcolor[HTML]{F59194}29.58 & \cellcolor[HTML]{F59194}.9816 & \cellcolor[HTML]{F59194}.0225 & 29.99 & .9176 & \cellcolor[HTML]{FAC791}.0789 \\
        Ours      & \cellcolor[HTML]{FAC791}21.09 & .9406 & \cellcolor[HTML]{FAC791}.0461 & \cellcolor[HTML]{FAC791}20.51 & \cellcolor[HTML]{FAC791}.9184 & \cellcolor[HTML]{FAC791}.0760 & 26.63 & .9714 & .0361 & \cellcolor[HTML]{FAC791}30.75 & \cellcolor[HTML]{F59194}.9281 & \cellcolor[HTML]{F59194}.0729 \\ \Xhline{1.2pt}
      & \multicolumn{3}{c}{\textbf{Mutant}} & \multicolumn{3}{c}{\textbf{Standup}}       & \multicolumn{3}{c}{\textbf{Hook}} & \multicolumn{3}{c}{\textbf{Average}}       \\ \cmidrule(lr){2-4} \cmidrule(lr){5-7} \cmidrule(lr){8-10} \cmidrule(lr){11-13} 
        \multirow{-2}{*}{Methods} & PSNR($\uparrow$)     & SSIM($\uparrow$)   & LPIPS($\downarrow$)   & PSNR($\uparrow$)    & SSIM($\uparrow$)    & LPIPS($\downarrow$)    & PSNR($\uparrow$)   & SSIM($\uparrow$)  & LPIPS($\downarrow$)  & PSNR($\uparrow$)     & SSIM($\uparrow$)    & LPIPS($\downarrow$)    \\ \toprule
        TiNeuVox-B    & 24.40                     & .9282 & .0700 & 21.77 & .9169 & .0927 & 21.05 & .8817 & .1033 & 22.83 & .9229 & .0886 \\
        4D-GS         & 24.61                     & .9269 & .0582 & 22.25 & .9140 & .0870 & \cellcolor[HTML]{F59194}23.93 & \cellcolor[HTML]{FAC791}.9042 & .0755 & 23.98 & .9305 & .0697 \\
        Deformable-GS & 24.32                     & .9300 & .0469 & 21.38 & .9133 & .0837 & 21.41 & .8872 & .0824 & 23.35 & .9285 & .0623 \\ \hline
        Ours-MLP      & \cellcolor[HTML]{FAC791}25.05                      & \cellcolor[HTML]{FAC791}.9359 & \cellcolor[HTML]{FAC791}.0409 & \cellcolor[HTML]{FAC791}23.04 & \cellcolor[HTML]{FAC791}.9250 & \cellcolor[HTML]{FAC791}.0700 & 22.6 & .8971 & \cellcolor[HTML]{FAC791}.0702 & \cellcolor[HTML]{FAC791}24.14 & \cellcolor[HTML]{FAC791}.9339 & \cellcolor[HTML]{FAC791}.0560 \\
        Ours      & \cellcolor[HTML]{F59194}28.16                      & \cellcolor[HTML]{F59194}.9560 & \cellcolor[HTML]{F59194}.0256 & \cellcolor[HTML]{F59194}25.96 & \cellcolor[HTML]{F59194}.9403 & \cellcolor[HTML]{F59194}.0481 & \cellcolor[HTML]{FAC791}23.42 & \cellcolor[HTML]{F59194}.9089 & \cellcolor[HTML]{F59194}.0573 & \cellcolor[HTML]{F59194}24.62 & \cellcolor[HTML]{F59194}.9387 & \cellcolor[HTML]{F59194}.0514 \\ \Xhline{1.2pt}
    \end{tabular}
    % \vspace{-4mm}
    \label{tab:D-NeRF_extrapolate}
\end{table*}

In this section, we first introduce our implementation details and the test datasets in Sec.~\ref{ssec:data}. Then we evaluate the dynamic scene rendering quality of our method in Sec.~\ref{ssec:rendering} and compare the prediction results with different baselines in Sec.~\ref{ssec:prediction}. Lastly, we perform ablation studies to analyze the effectiveness of our method in Sec.~\ref{ssec:ablation}.

\subsection{Datasets and Implementation Details}
\label{ssec:data}
We conduct evaluations on both synthetic and real datasets.

\paragraph{Synthetic Dataset.} 
The D-NeRF dataset \cite{D-NeRF} contains 8 dynamic objects, and
comprises 100-200 training images and 20 test images, with timestamps ranging from 0 to 1 for all images. We render images of this dataset at an 800 $\times$ 800 resolution. Following Deform-GS~\cite{Deformable-Gaussian}, we evaluate the rendering results with a black background and exclude the "\textit{Lego}" data because its test data and model did not align with the training data.

\paragraph{Real-World Dataset.} 
From the Hyper-NeRF real-world dataset~\cite{Hyper-NeRF},
we chose three scenes (\textit{cut-lemon, split-cookie, and chickchicken}) captured by a camera and one scene (\textit{3D-printer}) captured by two Google Pixel 3 phones.
Timestamps range from 0 to 1 for all images and rendering resolution is set to 960 $\times$ 540.
\if 0
\begin{itemize}
    \item \textbf{Synthetic Dataset.} We test our method on the D-NeRF dataset which contains 8 dynamic objects. This dataset comprises 100-200 training images and 20 test images, with timestamps ranging from 0 to 1 for all images. We render images of this dataset at an 800 $\times$ 800 resolution. Following the setup of Deform-GS~\cite{Deformable-Gaussian}, we evaluate the rendering results of all comparative methods with a black background and exclude the "\textit{Lego}" data because its test data and model did not align with the training data.
    \item \textbf{Real World Dataset.} We evaluate the performance of our model on the Hyper-NeRF real-world dataset~\cite{Hyper-NeRF}. We chose three scenes (\textit{cut-lemon, split-cookie, and chickchicken}) captured by a monocular camera and one scene (\textit{3D-printer}) captured by two Google Pixel 3 phones.
    Timestamps range from 0 to 1 for all images and rendering resolution is set to 960 $\times$ 540.
\end{itemize}
\fi

\paragraph{Implementation details.} 
We develop a composited training strategy, which learns scene hyper-canonical space, key points, and scene prediction in a three-step fashion. We train for 30k iterations in the first step. During the initial 1k iterations, we exclusively optimized the parameters of the 3D Gaussians for warm-up purposes. Subsequently, we jointly train for 27k iterations on the deformation MLP and the motion feature in Eq.~\ref{eq:deformable}. In the second step, we conduct training on the hash-encoding, the deformation MLP, and the position and motion feature of the key points for 10k iterations. Then in the third step, we train all the parameters together for 20k iterations using the synthetic dataset and 30k iterations for the real-world dataset. All our experiments are evaluated on an NVIDIA GeForce RTX 4090 GPU. 

\subsection{Comparison of Dynamic Scene Rendering Quality}
\label{ssec:rendering}
\begin{table*}[t]
\centering
\caption{Quantitative rendering results comparison on D-NeRF dataset. Best results are highlighted as \fs{\bf{first}},~\nd{\bf{second.}}}
% \vspace{-4mm}
\setlength{\extrarowheight}{2.5pt}
    \footnotesize
    \begin{tabular}{lcccccccccccc}
        \Xhline{1.2pt}
      & \multicolumn{3}{c}{\textbf{Trex}}        & \multicolumn{3}{c}{\textbf{Jumpingjacks}} & \multicolumn{3}{c}{\textbf{Bouncingballs}}    & \multicolumn{3}{c}{\textbf{Hellwarrior}} \\ \cmidrule(lr){2-4} \cmidrule(lr){5-7} \cmidrule(lr){8-10} \cmidrule(lr){11-13} 
        \multirow{-2}{*}{Method}  & PSNR($\uparrow$)     & SSIM($\uparrow$)   & LPIPS($\downarrow$)   & PSNR($\uparrow$)    & SSIM($\uparrow$)    & LPIPS($\downarrow$)    & PSNR($\uparrow$)   & SSIM($\uparrow$)  & LPIPS($\downarrow$)  & PSNR($\uparrow$)     & SSIM($\uparrow$)    & LPIPS($\downarrow$)    \\ \toprule
        TiNeuVox-B    & 31.24 & .9771 & .0326 & 34.29 & .9799 & .0360 & 35.00 & .9835 & .0391 & 39.20 & .9763 & .0508 \\
        4D-GS         & 33.60                      & .9863 & .0188 & 35.59 & .9844 & .0210 & 37.69 & .9919 & .0150 & 38.52 & .9754 & .0524 \\
        Deformable-GS & \cellcolor[HTML]{F59194}38.10 & \cellcolor[HTML]{F59194}.9933 & \cellcolor[HTML]{F59194}.0098 & \cellcolor[HTML]{FAC791}37.72 & \cellcolor[HTML]{FAC791}.9897 & \cellcolor[HTML]{FAC791}.0126 & \cellcolor[HTML]{FAC791}41.01 & \cellcolor[HTML]{FAC791}.9953 & \cellcolor[HTML]{FAC791}.0093 & \cellcolor[HTML]{FAC791}41.54 & \cellcolor[HTML]{FAC791}.9873 & \cellcolor[HTML]{FAC791}.0234 \\ \hline
        Ours          & \cellcolor[HTML]{FAC791}37.39                         & \cellcolor[HTML]{FAC791}.9926 & \cellcolor[HTML]{FAC791}.0110 & \cellcolor[HTML]{F59194}37.93 & \cellcolor[HTML]{F59194}.9906 & \cellcolor[HTML]{F59194}.0099 & \cellcolor[HTML]{F59194}41.57 & \cellcolor[HTML]{F59194}.9954 & \cellcolor[HTML]{F59194}.0086 & \cellcolor[HTML]{F59194}41.73 & \cellcolor[HTML]{F59194}.9874 & \cellcolor[HTML]{F59194}.0214 \\ \Xhline{1.2pt}
      & \multicolumn{3}{c}{\textbf{Mutant}} & \multicolumn{3}{c}{\textbf{Standup}}       & \multicolumn{3}{c}{\textbf{Hook}} & \multicolumn{3}{c}{\textbf{Average}}       \\ \cmidrule(lr){2-4} \cmidrule(lr){5-7} \cmidrule(lr){8-10} \cmidrule(lr){11-13} 
        \multirow{-2}{*}{Methods} & PSNR($\uparrow$)     & SSIM($\uparrow$)   & LPIPS($\downarrow$)   & PSNR($\uparrow$)    & SSIM($\uparrow$)    & LPIPS($\downarrow$)    & PSNR($\uparrow$)   & SSIM($\uparrow$)  & LPIPS($\downarrow$)  & PSNR($\uparrow$)     & SSIM($\uparrow$)    & LPIPS($\downarrow$)    \\ \toprule
        TiNeuVox-B    & 35.07                     & .9768 & .0307 & 38.11 & .9854 & .0208 & 33.34 & .9711 & .0458 & 35.18 & .9786 & .0365 \\
        4D-GS         & 38.80                     & .9857 & .0212 & 40.43 & .9890 & .0164 & 33.83 & .9728 & .0338 & 36.92 & .9836 & .0255 \\
        Deformable-GS & \cellcolor[HTML]{FAC791}42.63                     & \cellcolor[HTML]{FAC791}.9951 & \cellcolor[HTML]{FAC791}.0052 & \cellcolor[HTML]{FAC791}44.62 & \cellcolor[HTML]{FAC791}.9951 & \cellcolor[HTML]{FAC791}.0063 & \cellcolor[HTML]{FAC791}37.42 & \cellcolor[HTML]{FAC791}.9867 & \cellcolor[HTML]{FAC791}.0144 & \cellcolor[HTML]{FAC791}40.43 & \cellcolor[HTML]{FAC791}.9918 & \cellcolor[HTML]{FAC791}.0116 \\ \hline
        Ours          & \cellcolor[HTML]{F59194}42.90                      & \cellcolor[HTML]{F59194}.9954 & \cellcolor[HTML]{F59194}.0049 & \cellcolor[HTML]{F59194}45.09 & \cellcolor[HTML]{F59194}.9954 & \cellcolor[HTML]{F59194}.0057 & \cellcolor[HTML]{F59194}37.44 & \cellcolor[HTML]{F59194}.9868 & \cellcolor[HTML]{F59194}.0137 & \cellcolor[HTML]{F59194}40.58 & \cellcolor[HTML]{F59194}.9919 & \cellcolor[HTML]{F59194}.0107 \\ \Xhline{1.2pt}
    \end{tabular}
    % \vspace{-2mm}
    \label{tab:D-NeRF_interpolation}
\end{table*}
\begin{table*}[h]
\setlength{\extrarowheight}{2.5pt}
\caption{Quantitative results comparison with TiNeuVox~\cite{TiNeuVox}, HyperNeRF~\cite{Hyper-NeRF}, 4D-Gs~\cite{4D-Gaussians}, and Deform-GS~\cite{Deformable-Gaussian} on Hyper-NeRF real-dataset. Best results are highlighted as \fs{\bf{first}},\nd{\bf{second}},\third{\bf{third.}}}
% \vspace{-4mm}
\begin{center}
\footnotesize
\begin{tabular}{lcccccccccc}
\Xhline{1.2pt}
& \multicolumn{2}{c}{\textbf{CHICKEN}}        & \multicolumn{2}{c}{\textbf{CUT LEMON}} & \multicolumn{2}{c}{\textbf{SPLIT COOKIE}}    & \multicolumn{2}{c}{\textbf{3D PRINTER}} & \multicolumn{2}{c}{\multirow{2}{*}{\textbf{AVERAGE}}}  \\ 
& \multicolumn{2}{c}{\textbf{(113 images)}}        & \multicolumn{2}{c}{\textbf{(415 images)}} & \multicolumn{2}{c}{\textbf{(134 images)}}    & \multicolumn{2}{c}{\textbf{(207 images)}}  \\

\cmidrule(lr){2-3} \cmidrule(lr){4-5} \cmidrule(lr){6-7} \cmidrule(lr){8-9} \cmidrule(lr){10-11} 
\multirow{-3}{*}{Method}  & PSNR($\uparrow$)     & MS-SSIM($\uparrow$)   & PSNR($\uparrow$)    & MS-SSIM($\uparrow$)    & PSNR($\uparrow$)   & MS-SSIM($\uparrow$)   & PSNR($\uparrow$)     & MS-SSIM($\uparrow$)     & PSNR($\uparrow$)    & MS-SSIM($\uparrow$) \\ \toprule
TiNeuVox-B   & \cellcolor[HTML]{FAC791}27.7 &\cellcolor[HTML]{F59194} .951 & 28.6 &\cellcolor[HTML]{FAC791} .955 & 28.9 & .965 & \cellcolor[HTML]{F59194}22.8 & \cellcolor[HTML]{F59194}.839 & 27.2 & \cellcolor[HTML]{F59194}.928  \\
HyperNeRF      & \cellcolor[HTML]{F59194}28.7   & \cellcolor[HTML]{FAC791}.948 & \cellcolor[HTML]{F59194}31.8 & \cellcolor[HTML]{F59194}.956 & 30.9 & .967 & 20.0 & \cellcolor[HTML]{FAC791}.821 & \cellcolor[HTML]{FAC791}28.4 &  \cellcolor[HTML]{FAC791}.924 \\
4D-GS         & 26.9   & .911 & \cellcolor[HTML]{FFFF99}30.0 & .929 & \cellcolor[HTML]{FFFF99}32.5 & \cellcolor[HTML]{FFFF99}.975 & \cellcolor[HTML]{FFFF99}22.0 & .808 & \cellcolor[HTML]{FFFF99}28.1 &.905  \\
Deform-GS & 26.1 & .902 & 29.1 & .937 & \cellcolor[HTML]{FAC791}32.8 & \cellcolor[HTML]{FAC791}.981 & 20.3 & .756 & 27.2 & .896  \\ \hline
Ours          & \cellcolor[HTML]{FFFF99}27.1  & \cellcolor[HTML]{FFFF99}.920 & \cellcolor[HTML]{FAC791}31.1 & \cellcolor[HTML]{FFFF99}.952 & \cellcolor[HTML]{F59194}34.0 & \cellcolor[HTML]{F59194}.983 & \cellcolor[HTML]{FAC791}22.2 & \cellcolor[HTML]{FFFF99}.814 & \cellcolor[HTML]{F59194}28.9 & \cellcolor[HTML]{FFFF99}.920  \\ \Xhline{1.2pt}
  
\end{tabular}
\end{center}
% \vspace{-1mm}
\label{tab:Hyper_interpolation}
\end{table*}

\begin{figure*}[t]
  \centering
  \includegraphics[width=0.8\linewidth, trim={0 0 0 0}, clip]{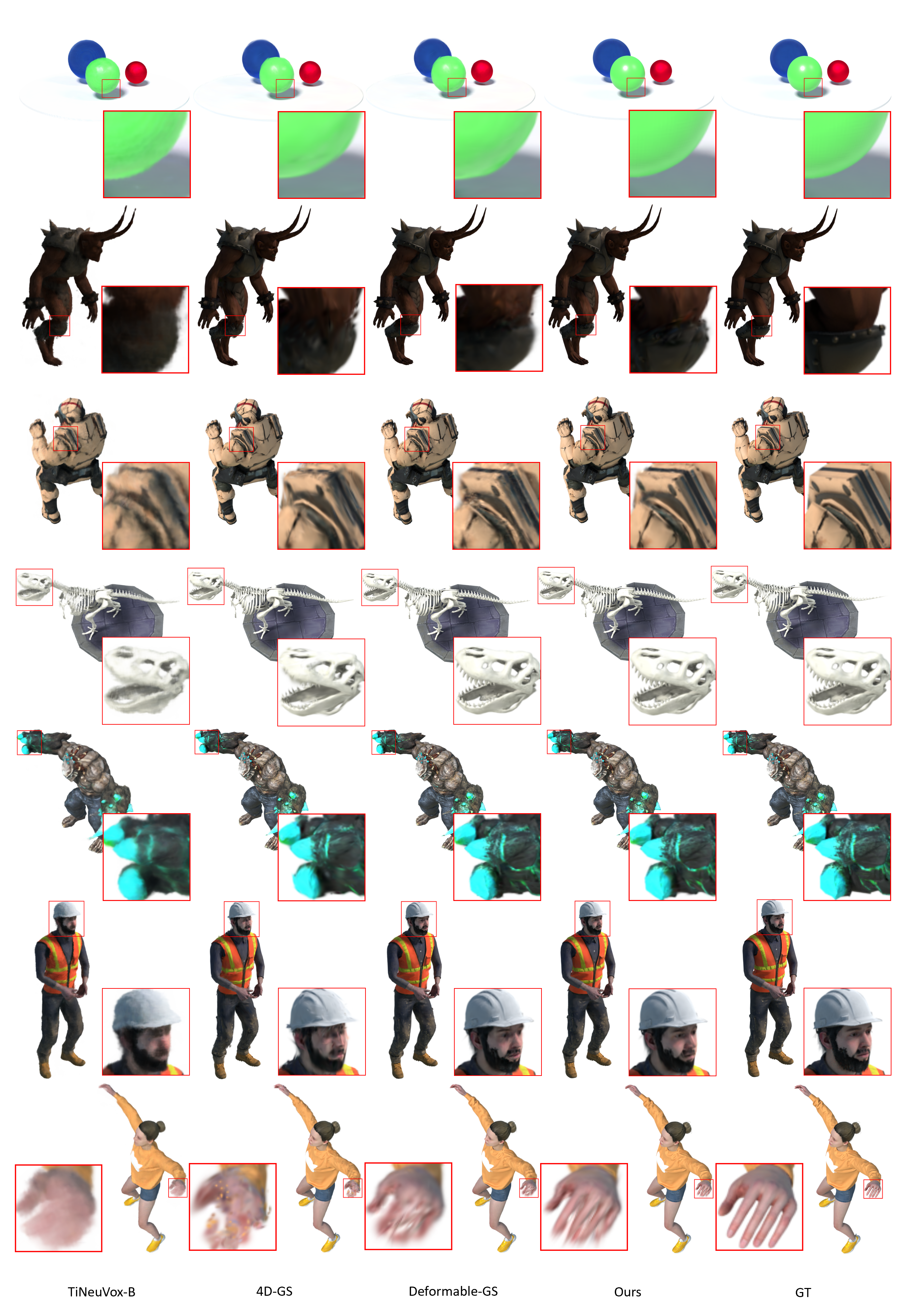}
  \caption{Qualitative results on synthetic dynamic scenes. We compare our methods with TiNeuVox~\cite{TiNeuVox}, 4D-GS~\cite{4D-Gaussians}, and Deform-GS~\cite{Deformable-Gaussian}.}
  \label{fig:D-NeRF_rendering}
\end{figure*}

We first compare our method with the current state-of-the-art dynamic scene reconstruction methods: TiNeuVox~\cite{TiNeuVox}, 4D-GS~\cite{4D-Gaussians} and Deform-GS~\cite{Deformable-Gaussian} on the synthetic dataset using the same data setting. The quantitative results are shown in Table~\ref{tab:D-NeRF_interpolation}. We present the PSNR/SSIM/LPIPS(VGG) values on this dataset. The results demonstrate that our method outperforms either existing NeRF-based or Gaussian-based methods. We also show the rendering results in Fig.~\ref{fig:D-NeRF_rendering}. It can be seen that our method achieves higher quality than other methods and reconstructs more details of dynamic scenes. 

\begin{figure*}[!t]
    \centering
    %Cookie
    \begin{subfigure}{0.16\linewidth}
        \centering
        \caption{TiNeuVox}
        \includegraphics[width=0.99\linewidth]{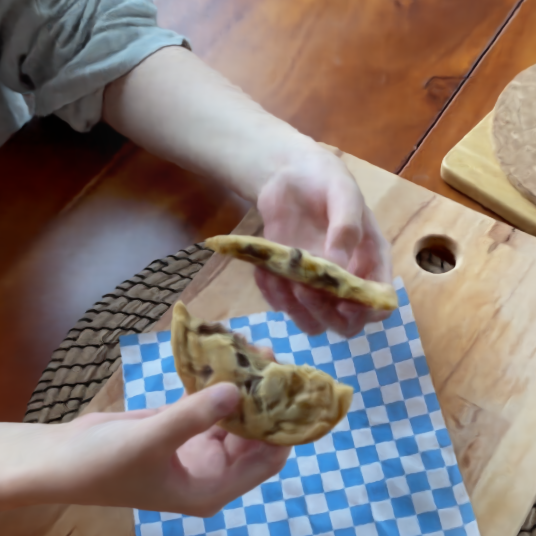}
    \end{subfigure}
    \begin{subfigure}{0.16\linewidth}
        \centering
        \caption{Hyper-NeRF}
        \includegraphics[width=0.99\linewidth]{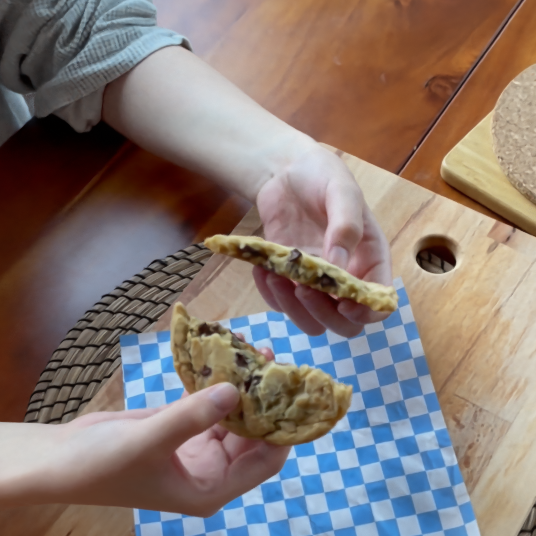}
    \end{subfigure}
    \begin{subfigure}{0.16\linewidth}
        \centering
        \caption{4D-GS}
        \includegraphics[width=0.99\linewidth]{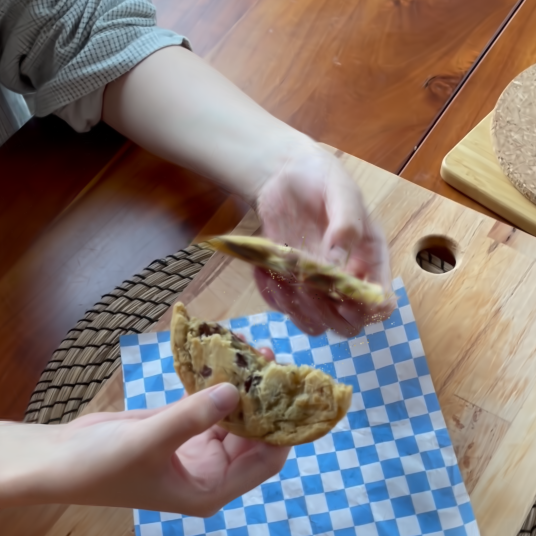}
    \end{subfigure}
    \begin{subfigure}{0.16\linewidth}
        \centering
        \caption{Deform-GS}
        \includegraphics[width=0.99\linewidth]{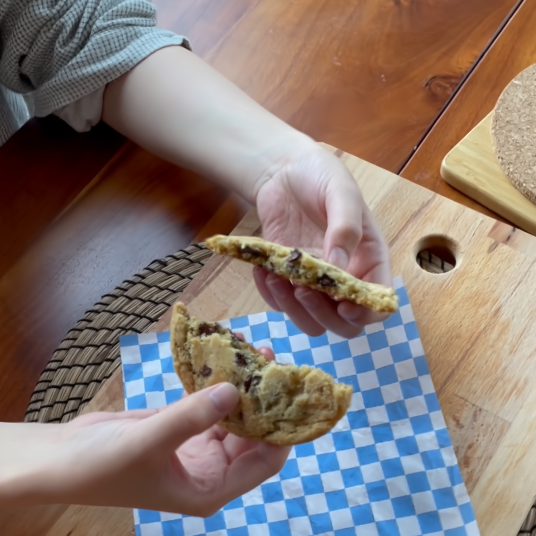}
    \end{subfigure}
    \begin{subfigure}{0.16\linewidth}
        \centering
        \caption{Ours}
        \includegraphics[width=0.99\linewidth]{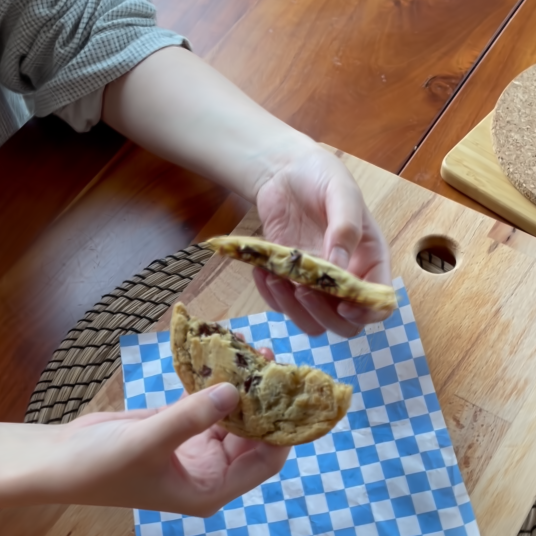}
    \end{subfigure}
    \begin{subfigure}{0.16\linewidth}
        \centering
        \caption{Ground Truth}
        \includegraphics[width=0.99\linewidth]{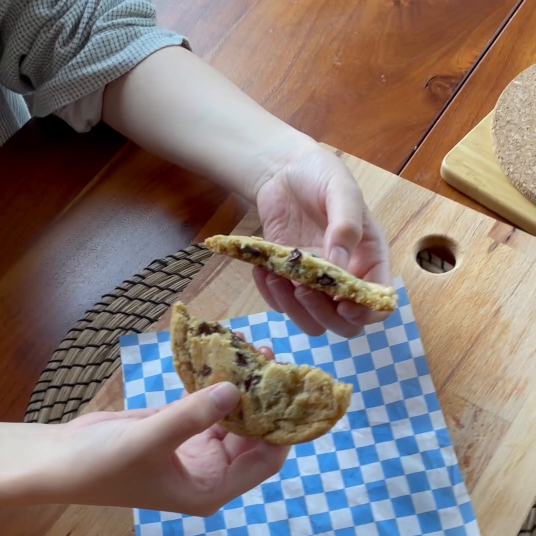}
    \end{subfigure}

    % Chicken
    \begin{subfigure}{0.16\linewidth}
        \centering
        \includegraphics[width=0.99\linewidth]{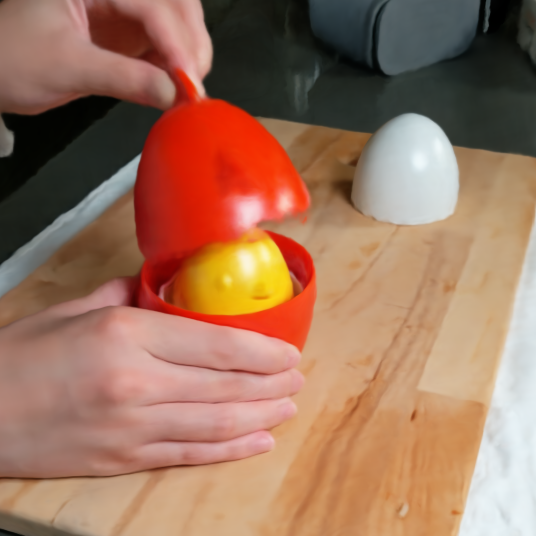}
    \end{subfigure}
    \begin{subfigure}{0.16\linewidth}
        \centering
        \includegraphics[width=0.99\linewidth]{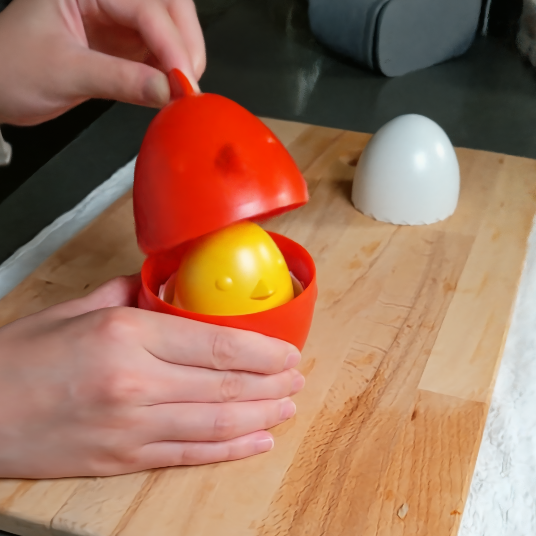}
    \end{subfigure}
    \begin{subfigure}{0.16\linewidth}
        \centering
        \includegraphics[width=0.99\linewidth]{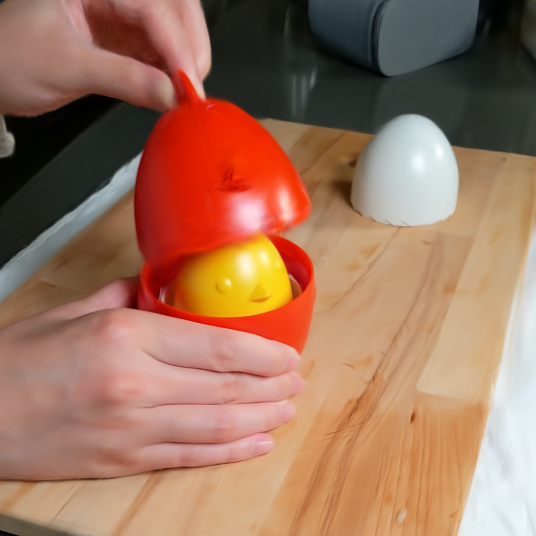}
    \end{subfigure}
    \begin{subfigure}{0.16\linewidth}
        \centering
        \includegraphics[width=0.99\linewidth]{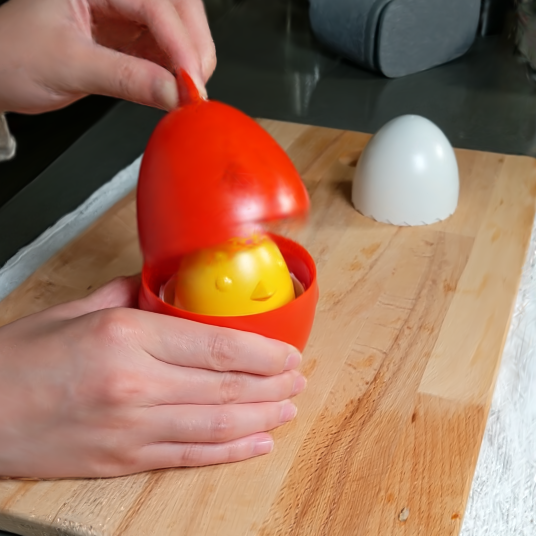}
    \end{subfigure}
    \begin{subfigure}{0.16\linewidth}
        \centering
        \includegraphics[width=0.99\linewidth]{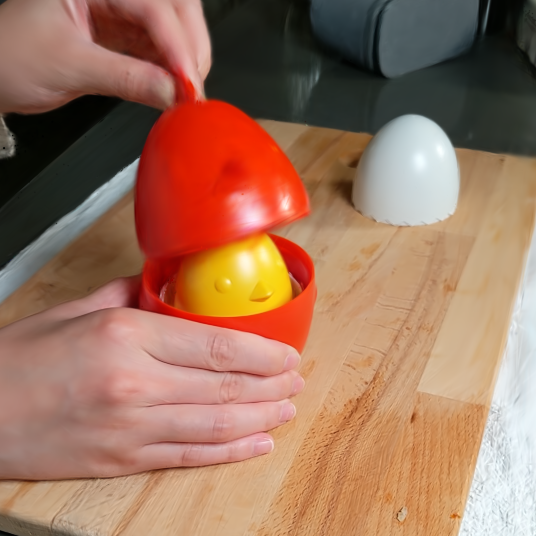}
    \end{subfigure}
    \begin{subfigure}{0.16\linewidth}
        \centering
        \includegraphics[width=0.99\linewidth]{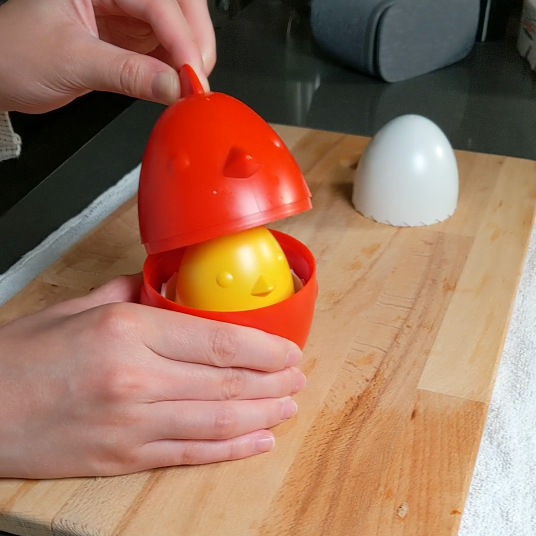}
    \end{subfigure}

    % Lemon
    \begin{subfigure}{0.16\linewidth}
        \centering
        \includegraphics[width=0.99\linewidth]{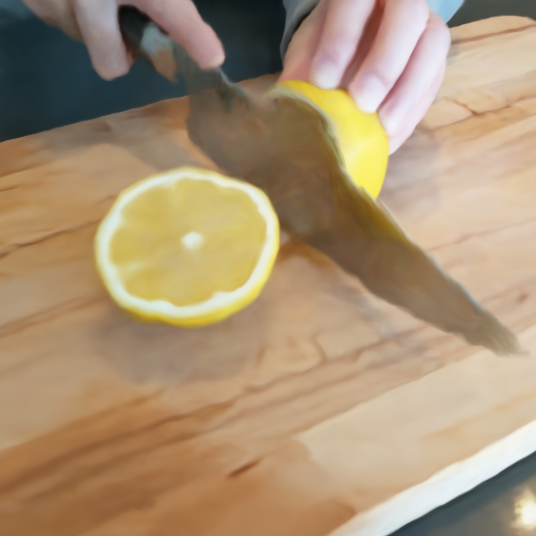}
    \end{subfigure}
    \begin{subfigure}{0.16\linewidth}
        \centering
        \includegraphics[width=0.99\linewidth]{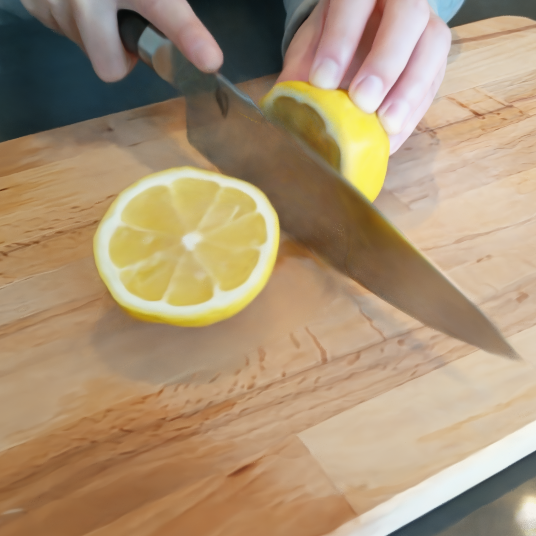}
    \end{subfigure}
    \begin{subfigure}{0.16\linewidth}
        \centering
        \includegraphics[width=0.99\linewidth]{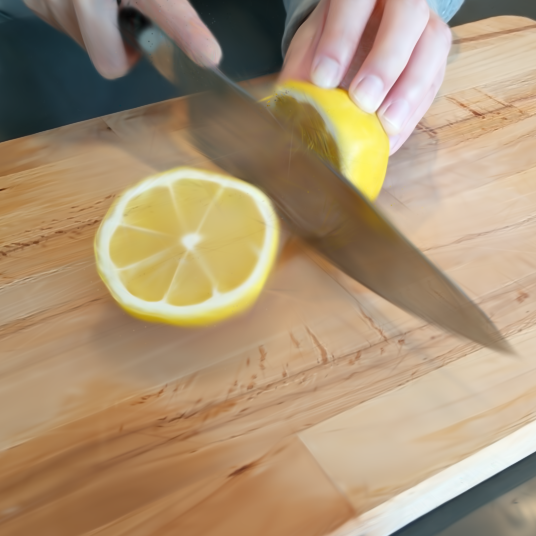}
    \end{subfigure}
    \begin{subfigure}{0.16\linewidth}
        \centering
        \includegraphics[width=0.99\linewidth]{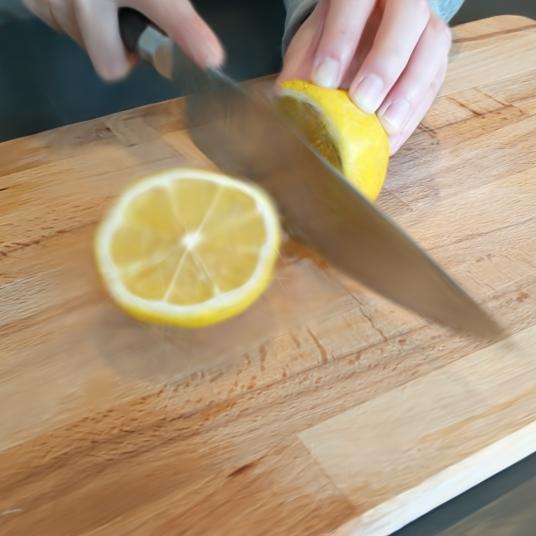}
    \end{subfigure}
    \begin{subfigure}{0.16\linewidth}
        \centering
        \includegraphics[width=0.99\linewidth]{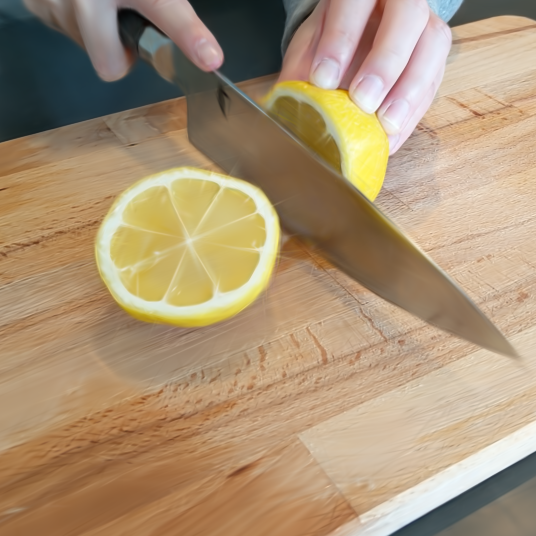}
    \end{subfigure}
    \begin{subfigure}{0.16\linewidth}
        \centering
        \includegraphics[width=0.99\linewidth]{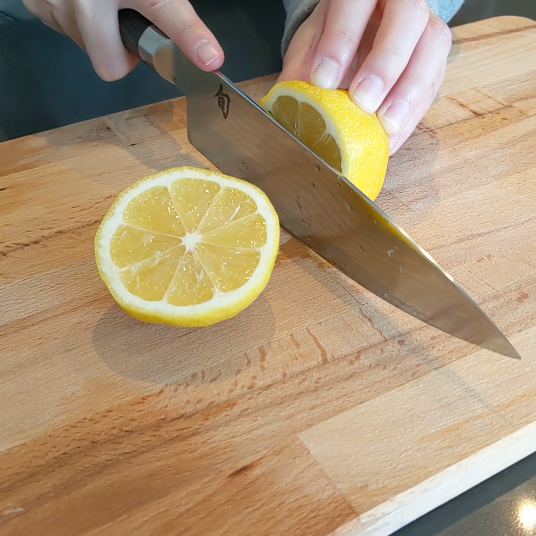}
    \end{subfigure}

    % \vspace{-3mm}
    % \captionsetup{labelsep=period, labelfont=bf}
    \caption{Qualitative results on real-world scenes. We compare our methods with TiNeuVox~\cite{TiNeuVox}, Hyper-NeRF~\cite{Hyper-NeRF}, 4D-GS~\cite{4D-Gaussians}, and Deform-GS~\cite{Deformable-Gaussian}.}
    % \vspace{-1mm}
    \label{fig:hyper_rendering}
\end{figure*}

We also evaluate our method on the real-world dataset following Hyper-NeRF's ~\cite{Hyper-NeRF} setting. We compare with four state-of-the-art methods and report the PSNR/MS-SSIM values in Table~\ref{tab:Hyper_interpolation}. As shown in Fig.~\ref{fig:hyper_rendering}, our method achieves better rendering quality on the real-world dataset. However, our method does not surpass the previous NeRF-based rendering~\cite{Hyper-NeRF, TiNeuVox} in quantitative results due to inaccurate ground truth camera poses and the misalignment of timestamps across all images in the real dataset. Despite this, ours still outperforms existing Gaussian-based methods~\cite{4D-Gaussians, Deformable-Gaussian}, which demonstrates the effectiveness of our approach.

\subsection{Comparison of Future Synthesis}
\label{ssec:prediction}
We now conduct the comparison on the future synthesis task in Table~\ref{tab:D-NeRF_extrapolate}. In this experiment, to synchronize the time intervals, we divide the original training data into new test and training sets. We utilize the data from the original training set with image timestamps less than 0.8s as the training data, and those greater than 0.8s as the test data. We directly input the time $t \sim [0.8-1.0]$ to each method and calculate the PSNR/SSIM/LPIPS(VGG) as presented in Table~\ref{tab:D-NeRF_extrapolate}. Additionally, we also compare with a couple of variants of GaussianPrediction: "Ours" is our full model which uses the GCN to predict the motion of key points (Sec.~\ref{ssec:gcn}), and "Ours-MLP" removes GCN and directly input the time $t$ to the deformable MLP in the second stage. 
Quantitative results demonstrate that our method achieves more realistic effects in future synthesis tasks, thanks to our key points distilled motion strategy. The experiments show that our GCN can learn the relationships between key points and predict more reasonable results which improves the rendering results of novel views in the future. We also show prediction quantitative evaluations of the real-world HyperNeRF dataset in Sec.~B of the supp. material. Note that in real-world scene-level datasets, the predicted results cannot be perfectly aligned with the ground truth images due to ill camera poses and inaccurate timestamps, which makes the quantitative comparison less meaningful than the qualitative comparison. 
Therefore, relying solely on quantitative evaluations to assess prediction performance is limited. Please refer to our video for more information. In short sequence prediction, our approach maintains better coherence, consistency, and reasonable.

\subsection{Ablation Studies}
\label{ssec:ablation}

\paragraph{Annealing noise.}
\label{sssec:ablation_mfeature}
\begin{table}[t!]
\centering
\caption{
%We perform 
Ablation studies of the annealing noise on the D-NeRF dataset.
}
% \vspace{-4mm}
\resizebox{1.0\linewidth}{!}{
\tabcolsep 14pt
\footnotesize
\begin{tabular}{lccc}
\toprule
\multicolumn{1}{c}{\multirow{2}{*}{Config.}} & \multicolumn{3}{c}{D-NeRF} \\ \cmidrule(lr){2-4} 
\multicolumn{1}{c}{} & \multicolumn{1}{l}{PSNR $\uparrow$} & \multicolumn{1}{l}{SSIM $\uparrow$} & \multicolumn{1}{l}{LPIPS $\downarrow$}  \\ \hline
w/o Annealing Noise & 39.71 & 0.991 & 0.012 \\
Full Model & \textbf{40.58} & \textbf{0.992} & \textbf{0.011}  \\
\bottomrule
\end{tabular}
}
% \vspace{-4mm}
% \vspace{0.2em}
\label{tab:ablation_noise}
\end{table}
\begin{table}[t!]
\centering
\caption{
%We perform 
Ablation study of each step discussed in Sec.~\ref{ssec:keypoints} on the real-world Hyper-NeRF dataset. Best results are highlighted as \fs{\bf{first}},~\nd{\bf{second.}}}
% \vspace{-4mm}
\label{tab:ablation}

\resizebox{1.0\linewidth}{!}{
\tabcolsep 14pt
\footnotesize
\begin{tabular}{lccc}
\toprule
\multicolumn{1}{c}{\multirow{2}{*}{Config.}} & \multicolumn{3}{c}{Hyper-NeRF} \\ \cmidrule(lr){2-4} 
\multicolumn{1}{c}{} & \multicolumn{1}{l}{PSNR $\uparrow$} & \multicolumn{1}{l}{SSIM $\uparrow$} & \multicolumn{1}{l}{LPIPS $\downarrow$}  \\ \hline
w/o Hyper Initialization & 27.9 & .911 & .232\\
w/o Adaptively Increasing & \cellcolor[HTML]{FAC791}28.6 & .917 & .225 \\
% w/o Hyper Near Search & 28.4 & .917 & .220\\
w/o Lifecycle & \cellcolor[HTML]{FAC791}28.6 & \cellcolor[HTML]{FAC791}.918 & \cellcolor[HTML]{FAC791}.218 \\
Full Model & \cellcolor[HTML]{F59194}28.9 & \cellcolor[HTML]{F59194}.920 & \cellcolor[HTML]{F59194}.184  \\
\bottomrule
\end{tabular}
}
% \vspace{-5mm}
% \vspace{0.2em}
\end{table}
We first inspect the effectiveness of the annealing noise during training hyper-canonical space (Sec.~\ref{ssec:canonical}). We show the location of 3D Gaussians in the canonical space in Fig.~\ref{fig:ablation_pcd}. When training with annealing noise, our model can preserve better geometry. The quantitative results in Table~\ref{tab:ablation_noise} indicate that training with annealing noise can improve the reconstruction results. This proves the necessity of annealing noise for training.

\paragraph{Initial key points in the hyper-canonical space.}
\label{sssec:ablation_clusterInit}
We also compared the impact of initializing key points in different spaces. As shown in Fig.~\ref{fig:ablation} and Table~\ref{tab:ablation}, directly initializing key points in 3D space results in significant blurring in regions with complex motion, leading to a decrease in quantitative values. However, our approach initializes key points in hyper-space, considering not only spatial information but also motion information, effectively improving rendering quality.

\paragraph{Adaptive increasing key points.}
To demonstrate the effectiveness of the adaptively increasing key points strategy, we compared the results of directly initializing $N_k$ key points with adaptively increasing $N_k - k_{init}$ key points in Table~\ref{tab:ablation}. The results demonstrate that our method excels in identifying regions that require additional key points, whereas direct initialization cannot adjust based on rendering outcomes.

\paragraph{Searching nearest key points in the hyper-canonical space.}
We then analyze the impact of searching the nearest key points in the hyper-canonical space. We choose a key point on the knife in \textit{Cut-Lemon} scene and then select all 3D Gaussians that are influenced by this key point using two different selecting strategies. As shown in Fig.~\ref{subfig:3D_select}, the key point on the knife is the nearest key point of 3D Gaussians on the lemon.
However, these two motions are entirely different yet influenced by the same key point, resulting in blurriness during rendering. Finding the nearest key point in the hyper-canonical space can prevent this issue. Fig.~\ref{fig:ablation} and Table~\ref{tab:ablation} further demonstrate the effectiveness of our approach.

\paragraph{Lifecycle opacity.}
\label{sssec:ablation_lifecycle}
We also study the effectiveness of lifecycle strategy in Fig.~\ref{fig:ablation} and Table~\ref{tab:ablation}. We can observe that when the lemon is sliced, some 3D Gaussians still have residues at the cut (within the red box), while the addition of the lifecycle strategy significantly eliminates these Gaussians, improving the rendering quality.

\section{Conclusion}
\label{sec:Conclusion}
In this paper, we present a novel framework, \ie, GaussianPrediction, for forecasting future scenarios in dynamic scenes. GaussianPrediction employs a 3D Gaussian canonical space with deformation modeling, coupled with a lifecycle property, to effectively represent changes in dynamic scenes. 
Additionally, a novel concentric motion distillation technique with key points is developed to simplify complex scene motion prediction with a Graph Convolutional Network.

Since GaussianPrediction learns to predict the dynamics of key points solely based on the input observations without any pre-training, it can only predict meaningful and short-term future scenarios. 
For long-term prediction, incorporating motion priors in our framework presents a promising direction for future research.

\begin{acks}
We would like to acknowledge the support from NSFC (No.~62102356), Ant Group, Information Technology Center, and State Key Lab of CAD\&CG, Zhejiang University. 
We also express our gratitude to all the anonymous reviewers for their professional and insightful comments. The authors from Zhejiang University are also affiliated with the State Key Lab of CAD\&CG. 
\end{acks}

\bibliographystyle{ACM-Reference-Format}
\bibliography{main}


\begin{thebibliography}{69}

%%% the Web address.

\ifx \showCODEN    \undefined \def \showCODEN     #1{\unskip}     \fi
\ifx \showDOI      \undefined \def \showDOI       #1{#1}\fi
\ifx \showISBNx    \undefined \def \showISBNx     #1{\unskip}     \fi
\ifx \showISBNxiii \undefined \def \showISBNxiii  #1{\unskip}     \fi
\ifx \showISSN     \undefined \def \showISSN      #1{\unskip}     \fi
\ifx \showLCCN     \undefined \def \showLCCN      #1{\unskip}     \fi
\ifx \shownote     \undefined \def \shownote      #1{#1}          \fi
\ifx \showarticletitle \undefined \def \showarticletitle #1{#1}   \fi
\ifx \showURL      \undefined \def \showURL       {\relax}        \fi
% invisible to TeX
\providecommand\bibfield[2]{#2}
\providecommand\bibinfo[2]{#2}
\providecommand\natexlab[1]{#1}
\providecommand\showeprint[2][]{arXiv:#2}

\bibitem[Alexanderson et~al\mbox{.}(2023)]%
        {alexanderson2023listen}
\bibfield{author}{\bibinfo{person}{Simon Alexanderson}, \bibinfo{person}{Rajmund Nagy}, \bibinfo{person}{Jonas Beskow}, {and} \bibinfo{person}{Gustav~Eje Henter}.} \bibinfo{year}{2023}\natexlab{}.
\newblock \showarticletitle{Listen, denoise, action! audio-driven motion synthesis with diffusion models}.
\newblock \bibinfo{journal}{\emph{ACM Transactions on Graphics (TOG)}} \bibinfo{volume}{42}, \bibinfo{number}{4} (\bibinfo{year}{2023}), \bibinfo{pages}{1--20}.
\newblock


\bibitem[Attal et~al\mbox{.}(2023)]%
        {attal2023hyperreel}
\bibfield{author}{\bibinfo{person}{Benjamin Attal}, \bibinfo{person}{Jia-Bin Huang}, \bibinfo{person}{Christian Richardt}, \bibinfo{person}{Michael Zollhoefer}, \bibinfo{person}{Johannes Kopf}, \bibinfo{person}{Matthew O’Toole}, {and} \bibinfo{person}{Changil Kim}.} \bibinfo{year}{2023}\natexlab{}.
\newblock \showarticletitle{HyperReel: High-fidelity 6-DoF video with ray-conditioned sampling}. In \bibinfo{booktitle}{\emph{Proceedings of the IEEE/CVF Conference on Computer Vision and Pattern Recognition}}. \bibinfo{pages}{16610--16620}.
\newblock


\bibitem[Babaeizadeh et~al\mbox{.}(2017)]%
        {babaeizadeh2017stochastic}
\bibfield{author}{\bibinfo{person}{Mohammad Babaeizadeh}, \bibinfo{person}{Chelsea Finn}, \bibinfo{person}{Dumitru Erhan}, \bibinfo{person}{Roy~H Campbell}, {and} \bibinfo{person}{Sergey Levine}.} \bibinfo{year}{2017}\natexlab{}.
\newblock \showarticletitle{Stochastic variational video prediction}.
\newblock \bibinfo{journal}{\emph{arXiv preprint arXiv:1710.11252}} (\bibinfo{year}{2017}).
\newblock


\bibitem[Bansal et~al\mbox{.}(2020)]%
        {bansal20204d}
\bibfield{author}{\bibinfo{person}{Aayush Bansal}, \bibinfo{person}{Minh Vo}, \bibinfo{person}{Yaser Sheikh}, \bibinfo{person}{Deva Ramanan}, {and} \bibinfo{person}{Srinivasa Narasimhan}.} \bibinfo{year}{2020}\natexlab{}.
\newblock \showarticletitle{4d visualization of dynamic events from unconstrained multi-view videos}. In \bibinfo{booktitle}{\emph{Proceedings of the IEEE/CVF Conference on Computer Vision and Pattern Recognition}}. \bibinfo{pages}{5366--5375}.
\newblock


\bibitem[Barquero et~al\mbox{.}(2023)]%
        {barquero2023belfusion}
\bibfield{author}{\bibinfo{person}{German Barquero}, \bibinfo{person}{Sergio Escalera}, {and} \bibinfo{person}{Cristina Palmero}.} \bibinfo{year}{2023}\natexlab{}.
\newblock \showarticletitle{Belfusion: Latent diffusion for behavior-driven human motion prediction}. In \bibinfo{booktitle}{\emph{Proceedings of the IEEE/CVF International Conference on Computer Vision}}. \bibinfo{pages}{2317--2327}.
\newblock


\bibitem[Barsoum et~al\mbox{.}(2018)]%
        {barsoum2018hp}
\bibfield{author}{\bibinfo{person}{Emad Barsoum}, \bibinfo{person}{John Kender}, {and} \bibinfo{person}{Zicheng Liu}.} \bibinfo{year}{2018}\natexlab{}.
\newblock \showarticletitle{Hp-gan: Probabilistic 3d human motion prediction via gan}. In \bibinfo{booktitle}{\emph{Proceedings of the IEEE conference on computer vision and pattern recognition workshops}}. \bibinfo{pages}{1418--1427}.
\newblock


\bibitem[Broxton et~al\mbox{.}(2020)]%
        {broxton2020immersive}
\bibfield{author}{\bibinfo{person}{Michael Broxton}, \bibinfo{person}{John Flynn}, \bibinfo{person}{Ryan Overbeck}, \bibinfo{person}{Daniel Erickson}, \bibinfo{person}{Peter Hedman}, \bibinfo{person}{Matthew Duvall}, \bibinfo{person}{Jason Dourgarian}, \bibinfo{person}{Jay Busch}, \bibinfo{person}{Matt Whalen}, {and} \bibinfo{person}{Paul Debevec}.} \bibinfo{year}{2020}\natexlab{}.
\newblock \showarticletitle{Immersive light field video with a layered mesh representation}.
\newblock \bibinfo{journal}{\emph{ACM Transactions on Graphics (TOG)}} \bibinfo{volume}{39}, \bibinfo{number}{4} (\bibinfo{year}{2020}), \bibinfo{pages}{86--1}.
\newblock


\bibitem[Cao and Johnson(2023)]%
        {cao2023hexplane}
\bibfield{author}{\bibinfo{person}{Ang Cao} {and} \bibinfo{person}{Justin Johnson}.} \bibinfo{year}{2023}\natexlab{}.
\newblock \showarticletitle{Hexplane: A fast representation for dynamic scenes}. In \bibinfo{booktitle}{\emph{Proceedings of the IEEE/CVF Conference on Computer Vision and Pattern Recognition}}. \bibinfo{pages}{130--141}.
\newblock


\bibitem[Corona et~al\mbox{.}(2020)]%
        {corona2020context}
\bibfield{author}{\bibinfo{person}{Enric Corona}, \bibinfo{person}{Albert Pumarola}, \bibinfo{person}{Guillem Alenya}, {and} \bibinfo{person}{Francesc Moreno-Noguer}.} \bibinfo{year}{2020}\natexlab{}.
\newblock \showarticletitle{Context-aware human motion prediction}. In \bibinfo{booktitle}{\emph{Proceedings of the IEEE/CVF Conference on Computer Vision and Pattern Recognition}}. \bibinfo{pages}{6992--7001}.
\newblock


\bibitem[Denton and Fergus(2018)]%
        {denton2018stochastic}
\bibfield{author}{\bibinfo{person}{Emily Denton} {and} \bibinfo{person}{Rob Fergus}.} \bibinfo{year}{2018}\natexlab{}.
\newblock \showarticletitle{Stochastic video generation with a learned prior}. In \bibinfo{booktitle}{\emph{International conference on machine learning}}. PMLR, \bibinfo{pages}{1174--1183}.
\newblock


\bibitem[Dou et~al\mbox{.}(2016)]%
        {dou2016fusion4d}
\bibfield{author}{\bibinfo{person}{Mingsong Dou}, \bibinfo{person}{Sameh Khamis}, \bibinfo{person}{Yury Degtyarev}, \bibinfo{person}{Philip Davidson}, \bibinfo{person}{Sean~Ryan Fanello}, \bibinfo{person}{Adarsh Kowdle}, \bibinfo{person}{Sergio~Orts Escolano}, \bibinfo{person}{Christoph Rhemann}, \bibinfo{person}{David Kim}, \bibinfo{person}{Jonathan Taylor}, {et~al\mbox{.}}} \bibinfo{year}{2016}\natexlab{}.
\newblock \showarticletitle{Fusion4d: Real-time performance capture of challenging scenes}.
\newblock \bibinfo{journal}{\emph{ACM Transactions on Graphics (ToG)}} \bibinfo{volume}{35}, \bibinfo{number}{4} (\bibinfo{year}{2016}), \bibinfo{pages}{1--13}.
\newblock


\bibitem[Du et~al\mbox{.}(2021)]%
        {du2021neural}
\bibfield{author}{\bibinfo{person}{Yilun Du}, \bibinfo{person}{Yinan Zhang}, \bibinfo{person}{Hong-Xing Yu}, \bibinfo{person}{Joshua~B Tenenbaum}, {and} \bibinfo{person}{Jiajun Wu}.} \bibinfo{year}{2021}\natexlab{}.
\newblock \showarticletitle{Neural radiance flow for 4d view synthesis and video processing}. In \bibinfo{booktitle}{\emph{2021 IEEE/CVF International Conference on Computer Vision (ICCV)}}. IEEE Computer Society, \bibinfo{pages}{14304--14314}.
\newblock


\bibitem[Eldar et~al\mbox{.}(1997)]%
        {FPS}
\bibfield{author}{\bibinfo{person}{Yuval Eldar}, \bibinfo{person}{Michael Lindenbaum}, \bibinfo{person}{Moshe Porat}, {and} \bibinfo{person}{Yehoshua~Y Zeevi}.} \bibinfo{year}{1997}\natexlab{}.
\newblock \showarticletitle{The farthest point strategy for progressive image sampling}.
\newblock \bibinfo{journal}{\emph{IEEE Transactions on Image Processing}} \bibinfo{volume}{6}, \bibinfo{number}{9} (\bibinfo{year}{1997}), \bibinfo{pages}{1305--1315}.
\newblock


\bibitem[Fang et~al\mbox{.}(2022a)]%
        {fang2022fast}
\bibfield{author}{\bibinfo{person}{Jiemin Fang}, \bibinfo{person}{Taoran Yi}, \bibinfo{person}{Xinggang Wang}, \bibinfo{person}{Lingxi Xie}, \bibinfo{person}{Xiaopeng Zhang}, \bibinfo{person}{Wenyu Liu}, \bibinfo{person}{Matthias Nie{\ss}ner}, {and} \bibinfo{person}{Qi Tian}.} \bibinfo{year}{2022}\natexlab{a}.
\newblock \showarticletitle{Fast dynamic radiance fields with time-aware neural voxels}. In \bibinfo{booktitle}{\emph{SIGGRAPH Asia 2022 Conference Papers}}. \bibinfo{pages}{1--9}.
\newblock


\bibitem[Fang et~al\mbox{.}(2022b)]%
        {TiNeuVox}
\bibfield{author}{\bibinfo{person}{Jiemin Fang}, \bibinfo{person}{Taoran Yi}, \bibinfo{person}{Xinggang Wang}, \bibinfo{person}{Lingxi Xie}, \bibinfo{person}{Xiaopeng Zhang}, \bibinfo{person}{Wenyu Liu}, \bibinfo{person}{Matthias Nie{\ss}ner}, {and} \bibinfo{person}{Qi Tian}.} \bibinfo{year}{2022}\natexlab{b}.
\newblock \showarticletitle{Fast dynamic radiance fields with time-aware neural voxels}. In \bibinfo{booktitle}{\emph{SIGGRAPH Asia 2022 Conference Papers}}. \bibinfo{pages}{1--9}.
\newblock


\bibitem[Fridovich-Keil et~al\mbox{.}(2023)]%
        {fridovich2023k}
\bibfield{author}{\bibinfo{person}{Sara Fridovich-Keil}, \bibinfo{person}{Giacomo Meanti}, \bibinfo{person}{Frederik~Rahb{\ae}k Warburg}, \bibinfo{person}{Benjamin Recht}, {and} \bibinfo{person}{Angjoo Kanazawa}.} \bibinfo{year}{2023}\natexlab{}.
\newblock \showarticletitle{K-planes: Explicit radiance fields in space, time, and appearance}. In \bibinfo{booktitle}{\emph{Proceedings of the IEEE/CVF Conference on Computer Vision and Pattern Recognition}}. \bibinfo{pages}{12479--12488}.
\newblock


\bibitem[Gao et~al\mbox{.}(2021)]%
        {gao2021dynamic}
\bibfield{author}{\bibinfo{person}{Chen Gao}, \bibinfo{person}{Ayush Saraf}, \bibinfo{person}{Johannes Kopf}, {and} \bibinfo{person}{Jia-Bin Huang}.} \bibinfo{year}{2021}\natexlab{}.
\newblock \showarticletitle{Dynamic view synthesis from dynamic monocular video}. In \bibinfo{booktitle}{\emph{Proceedings of the IEEE/CVF International Conference on Computer Vision}}. \bibinfo{pages}{5712--5721}.
\newblock


\bibitem[Geng et~al\mbox{.}(2023)]%
        {geng2023learning}
\bibfield{author}{\bibinfo{person}{Chen Geng}, \bibinfo{person}{Sida Peng}, \bibinfo{person}{Zhen Xu}, \bibinfo{person}{Hujun Bao}, {and} \bibinfo{person}{Xiaowei Zhou}.} \bibinfo{year}{2023}\natexlab{}.
\newblock \showarticletitle{Learning neural volumetric representations of dynamic humans in minutes}. In \bibinfo{booktitle}{\emph{Proceedings of the IEEE/CVF Conference on Computer Vision and Pattern Recognition}}. \bibinfo{pages}{8759--8770}.
\newblock


\bibitem[Girdhar and Grauman(2021)]%
        {girdhar2021anticipative}
\bibfield{author}{\bibinfo{person}{Rohit Girdhar} {and} \bibinfo{person}{Kristen Grauman}.} \bibinfo{year}{2021}\natexlab{}.
\newblock \showarticletitle{Anticipative video transformer}. In \bibinfo{booktitle}{\emph{Proceedings of the IEEE/CVF international conference on computer vision}}. \bibinfo{pages}{13505--13515}.
\newblock


\bibitem[H{\"o}ppe et~al\mbox{.}(2022)]%
        {hoppe2022diffusion}
\bibfield{author}{\bibinfo{person}{Tobias H{\"o}ppe}, \bibinfo{person}{Arash Mehrjou}, \bibinfo{person}{Stefan Bauer}, \bibinfo{person}{Didrik Nielsen}, {and} \bibinfo{person}{Andrea Dittadi}.} \bibinfo{year}{2022}\natexlab{}.
\newblock \showarticletitle{Diffusion models for video prediction and infilling}.
\newblock \bibinfo{journal}{\emph{arXiv preprint arXiv:2206.07696}} (\bibinfo{year}{2022}).
\newblock


\bibitem[Kania et~al\mbox{.}(2022)]%
        {kania2022conerf}
\bibfield{author}{\bibinfo{person}{Kacper Kania}, \bibinfo{person}{Kwang~Moo Yi}, \bibinfo{person}{Marek Kowalski}, \bibinfo{person}{Tomasz Trzci{\'n}ski}, {and} \bibinfo{person}{Andrea Tagliasacchi}.} \bibinfo{year}{2022}\natexlab{}.
\newblock \showarticletitle{Conerf: Controllable neural radiance fields}. In \bibinfo{booktitle}{\emph{Proceedings of the IEEE/CVF Conference on Computer Vision and Pattern Recognition}}. \bibinfo{pages}{18623--18632}.
\newblock


\bibitem[Kerbl et~al\mbox{.}(2023)]%
        {3DGaussian}
\bibfield{author}{\bibinfo{person}{Bernhard Kerbl}, \bibinfo{person}{Georgios Kopanas}, \bibinfo{person}{Thomas Leimk{\"u}hler}, {and} \bibinfo{person}{George Drettakis}.} \bibinfo{year}{2023}\natexlab{}.
\newblock \showarticletitle{3D Gaussian Splatting for Real-Time Radiance Field Rendering}.
\newblock \bibinfo{journal}{\emph{ACM Transactions on Graphics}} \bibinfo{volume}{42}, \bibinfo{number}{4} (\bibinfo{year}{2023}).
\newblock


\bibitem[Kwon and Park(2019)]%
        {kwon2019predicting}
\bibfield{author}{\bibinfo{person}{Yong-Hoon Kwon} {and} \bibinfo{person}{Min-Gyu Park}.} \bibinfo{year}{2019}\natexlab{}.
\newblock \showarticletitle{Predicting future frames using retrospective cycle gan}. In \bibinfo{booktitle}{\emph{Proceedings of the IEEE/CVF Conference on Computer Vision and Pattern Recognition}}. \bibinfo{pages}{1811--1820}.
\newblock


\bibitem[Li et~al\mbox{.}(2022)]%
        {li2022neural}
\bibfield{author}{\bibinfo{person}{Tianye Li}, \bibinfo{person}{Mira Slavcheva}, \bibinfo{person}{Michael Zollhoefer}, \bibinfo{person}{Simon Green}, \bibinfo{person}{Christoph Lassner}, \bibinfo{person}{Changil Kim}, \bibinfo{person}{Tanner Schmidt}, \bibinfo{person}{Steven Lovegrove}, \bibinfo{person}{Michael Goesele}, \bibinfo{person}{Richard Newcombe}, {et~al\mbox{.}}} \bibinfo{year}{2022}\natexlab{}.
\newblock \showarticletitle{Neural 3d video synthesis from multi-view video}. In \bibinfo{booktitle}{\emph{Proceedings of the IEEE/CVF Conference on Computer Vision and Pattern Recognition}}. \bibinfo{pages}{5521--5531}.
\newblock


\bibitem[Li et~al\mbox{.}(2021)]%
        {li2021neural}
\bibfield{author}{\bibinfo{person}{Zhengqi Li}, \bibinfo{person}{Simon Niklaus}, \bibinfo{person}{Noah Snavely}, {and} \bibinfo{person}{Oliver Wang}.} \bibinfo{year}{2021}\natexlab{}.
\newblock \showarticletitle{Neural scene flow fields for space-time view synthesis of dynamic scenes}. In \bibinfo{booktitle}{\emph{Proceedings of the IEEE/CVF Conference on Computer Vision and Pattern Recognition}}. \bibinfo{pages}{6498--6508}.
\newblock


\bibitem[Li et~al\mbox{.}(2023)]%
        {li2023dynibar}
\bibfield{author}{\bibinfo{person}{Zhengqi Li}, \bibinfo{person}{Qianqian Wang}, \bibinfo{person}{Forrester Cole}, \bibinfo{person}{Richard Tucker}, {and} \bibinfo{person}{Noah Snavely}.} \bibinfo{year}{2023}\natexlab{}.
\newblock \showarticletitle{Dynibar: Neural dynamic image-based rendering}. In \bibinfo{booktitle}{\emph{Proceedings of the IEEE/CVF Conference on Computer Vision and Pattern Recognition}}. \bibinfo{pages}{4273--4284}.
\newblock


\bibitem[Lin et~al\mbox{.}(2023)]%
        {lin2023im4d}
\bibfield{author}{\bibinfo{person}{Haotong Lin}, \bibinfo{person}{Sida Peng}, \bibinfo{person}{Zhen Xu}, \bibinfo{person}{Tao Xie}, \bibinfo{person}{Xingyi He}, \bibinfo{person}{Hujun Bao}, {and} \bibinfo{person}{Xiaowei Zhou}.} \bibinfo{year}{2023}\natexlab{}.
\newblock \showarticletitle{Im4D: High-Fidelity and Real-Time Novel View Synthesis for Dynamic Scenes}.
\newblock \bibinfo{journal}{\emph{arXiv preprint arXiv:2310.08585}} (\bibinfo{year}{2023}).
\newblock


\bibitem[Lin et~al\mbox{.}(2022)]%
        {lin2022efficient}
\bibfield{author}{\bibinfo{person}{Haotong Lin}, \bibinfo{person}{Sida Peng}, \bibinfo{person}{Zhen Xu}, \bibinfo{person}{Yunzhi Yan}, \bibinfo{person}{Qing Shuai}, \bibinfo{person}{Hujun Bao}, {and} \bibinfo{person}{Xiaowei Zhou}.} \bibinfo{year}{2022}\natexlab{}.
\newblock \showarticletitle{Efficient neural radiance fields for interactive free-viewpoint video}. In \bibinfo{booktitle}{\emph{SIGGRAPH Asia 2022 Conference Papers}}. \bibinfo{pages}{1--9}.
\newblock


\bibitem[Liu et~al\mbox{.}(2023)]%
        {Robust-dynamic-NeRF}
\bibfield{author}{\bibinfo{person}{Yu-Lun Liu}, \bibinfo{person}{Chen Gao}, \bibinfo{person}{Andreas Meuleman}, \bibinfo{person}{Hung-Yu Tseng}, \bibinfo{person}{Ayush Saraf}, \bibinfo{person}{Changil Kim}, \bibinfo{person}{Yung-Yu Chuang}, \bibinfo{person}{Johannes Kopf}, {and} \bibinfo{person}{Jia-Bin Huang}.} \bibinfo{year}{2023}\natexlab{}.
\newblock \showarticletitle{Robust dynamic radiance fields}. In \bibinfo{booktitle}{\emph{Proceedings of the IEEE/CVF Conference on Computer Vision and Pattern Recognition}}. \bibinfo{pages}{13--23}.
\newblock


\bibitem[Lombardi et~al\mbox{.}(2019)]%
        {lombardi2019neural}
\bibfield{author}{\bibinfo{person}{Stephen Lombardi}, \bibinfo{person}{Tomas Simon}, \bibinfo{person}{Jason Saragih}, \bibinfo{person}{Gabriel Schwartz}, \bibinfo{person}{Andreas Lehrmann}, {and} \bibinfo{person}{Yaser Sheikh}.} \bibinfo{year}{2019}\natexlab{}.
\newblock \showarticletitle{Neural volumes: Learning dynamic renderable volumes from images}.
\newblock \bibinfo{journal}{\emph{arXiv preprint arXiv:1906.07751}} (\bibinfo{year}{2019}).
\newblock


\bibitem[Lombardi et~al\mbox{.}(2021)]%
        {lombardi2021mixture}
\bibfield{author}{\bibinfo{person}{Stephen Lombardi}, \bibinfo{person}{Tomas Simon}, \bibinfo{person}{Gabriel Schwartz}, \bibinfo{person}{Michael Zollhoefer}, \bibinfo{person}{Yaser Sheikh}, {and} \bibinfo{person}{Jason Saragih}.} \bibinfo{year}{2021}\natexlab{}.
\newblock \showarticletitle{Mixture of volumetric primitives for efficient neural rendering}.
\newblock \bibinfo{journal}{\emph{ACM Transactions on Graphics (ToG)}} \bibinfo{volume}{40}, \bibinfo{number}{4} (\bibinfo{year}{2021}), \bibinfo{pages}{1--13}.
\newblock


\bibitem[Lu et~al\mbox{.}(2017)]%
        {lu2017flexible}
\bibfield{author}{\bibinfo{person}{Chaochao Lu}, \bibinfo{person}{Michael Hirsch}, {and} \bibinfo{person}{Bernhard Scholkopf}.} \bibinfo{year}{2017}\natexlab{}.
\newblock \showarticletitle{Flexible spatio-temporal networks for video prediction}. In \bibinfo{booktitle}{\emph{Proceedings of the IEEE Conference on Computer Vision and Pattern Recognition}}. \bibinfo{pages}{6523--6531}.
\newblock


\bibitem[Luiten et~al\mbox{.}(2023)]%
        {luiten2023dynamic}
\bibfield{author}{\bibinfo{person}{Jonathon Luiten}, \bibinfo{person}{Georgios Kopanas}, \bibinfo{person}{Bastian Leibe}, {and} \bibinfo{person}{Deva Ramanan}.} \bibinfo{year}{2023}\natexlab{}.
\newblock \showarticletitle{Dynamic 3d gaussians: Tracking by persistent dynamic view synthesis}.
\newblock \bibinfo{journal}{\emph{arXiv preprint arXiv:2308.09713}} (\bibinfo{year}{2023}).
\newblock


\bibitem[Mao et~al\mbox{.}(2019)]%
        {mao2019learning}
\bibfield{author}{\bibinfo{person}{Wei Mao}, \bibinfo{person}{Miaomiao Liu}, \bibinfo{person}{Mathieu Salzmann}, {and} \bibinfo{person}{Hongdong Li}.} \bibinfo{year}{2019}\natexlab{}.
\newblock \showarticletitle{Learning trajectory dependencies for human motion prediction}. In \bibinfo{booktitle}{\emph{Proceedings of the IEEE/CVF international conference on computer vision}}. \bibinfo{pages}{9489--9497}.
\newblock


\bibitem[Martinez et~al\mbox{.}(2017)]%
        {martinez2017human}
\bibfield{author}{\bibinfo{person}{Julieta Martinez}, \bibinfo{person}{Michael~J Black}, {and} \bibinfo{person}{Javier Romero}.} \bibinfo{year}{2017}\natexlab{}.
\newblock \showarticletitle{On human motion prediction using recurrent neural networks}. In \bibinfo{booktitle}{\emph{Proceedings of the IEEE conference on computer vision and pattern recognition}}. \bibinfo{pages}{2891--2900}.
\newblock


\bibitem[Mersch et~al\mbox{.}(2022)]%
        {mersch2022self}
\bibfield{author}{\bibinfo{person}{Benedikt Mersch}, \bibinfo{person}{Xieyuanli Chen}, \bibinfo{person}{Jens Behley}, {and} \bibinfo{person}{Cyrill Stachniss}.} \bibinfo{year}{2022}\natexlab{}.
\newblock \showarticletitle{Self-supervised point cloud prediction using 3d spatio-temporal convolutional networks}. In \bibinfo{booktitle}{\emph{Conference on Robot Learning}}. PMLR, \bibinfo{pages}{1444--1454}.
\newblock


\bibitem[Mildenhall et~al\mbox{.}(2021)]%
        {mildenhall2021nerf}
\bibfield{author}{\bibinfo{person}{Ben Mildenhall}, \bibinfo{person}{Pratul~P Srinivasan}, \bibinfo{person}{Matthew Tancik}, \bibinfo{person}{Jonathan~T Barron}, \bibinfo{person}{Ravi Ramamoorthi}, {and} \bibinfo{person}{Ren Ng}.} \bibinfo{year}{2021}\natexlab{}.
\newblock \showarticletitle{Nerf: Representing scenes as neural radiance fields for view synthesis}.
\newblock \bibinfo{journal}{\emph{Commun. ACM}} \bibinfo{volume}{65}, \bibinfo{number}{1} (\bibinfo{year}{2021}), \bibinfo{pages}{99--106}.
\newblock


\bibitem[M{\"u}ller et~al\mbox{.}(2022)]%
        {Instant-ngp}
\bibfield{author}{\bibinfo{person}{Thomas M{\"u}ller}, \bibinfo{person}{Alex Evans}, \bibinfo{person}{Christoph Schied}, {and} \bibinfo{person}{Alexander Keller}.} \bibinfo{year}{2022}\natexlab{}.
\newblock \showarticletitle{Instant neural graphics primitives with a multiresolution hash encoding}.
\newblock \bibinfo{journal}{\emph{ACM Transactions on Graphics (ToG)}} \bibinfo{volume}{41}, \bibinfo{number}{4} (\bibinfo{year}{2022}), \bibinfo{pages}{1--15}.
\newblock


\bibitem[Neimark et~al\mbox{.}(2021)]%
        {neimark2021video}
\bibfield{author}{\bibinfo{person}{Daniel Neimark}, \bibinfo{person}{Omri Bar}, \bibinfo{person}{Maya Zohar}, {and} \bibinfo{person}{Dotan Asselmann}.} \bibinfo{year}{2021}\natexlab{}.
\newblock \showarticletitle{Video transformer network}. In \bibinfo{booktitle}{\emph{Proceedings of the IEEE/CVF international conference on computer vision}}. \bibinfo{pages}{3163--3172}.
\newblock


\bibitem[Newcombe et~al\mbox{.}(2015)]%
        {newcombe2015dynamicfusion}
\bibfield{author}{\bibinfo{person}{Richard~A Newcombe}, \bibinfo{person}{Dieter Fox}, {and} \bibinfo{person}{Steven~M Seitz}.} \bibinfo{year}{2015}\natexlab{}.
\newblock \showarticletitle{Dynamicfusion: Reconstruction and tracking of non-rigid scenes in real-time}. In \bibinfo{booktitle}{\emph{Proceedings of the IEEE conference on computer vision and pattern recognition}}. \bibinfo{pages}{343--352}.
\newblock


\bibitem[Oprea et~al\mbox{.}(2020)]%
        {oprea2020review}
\bibfield{author}{\bibinfo{person}{Sergiu Oprea}, \bibinfo{person}{Pablo Martinez-Gonzalez}, \bibinfo{person}{Alberto Garcia-Garcia}, \bibinfo{person}{John~Alejandro Castro-Vargas}, \bibinfo{person}{Sergio Orts-Escolano}, \bibinfo{person}{Jose Garcia-Rodriguez}, {and} \bibinfo{person}{Antonis Argyros}.} \bibinfo{year}{2020}\natexlab{}.
\newblock \showarticletitle{A review on deep learning techniques for video prediction}.
\newblock \bibinfo{journal}{\emph{IEEE Transactions on Pattern Analysis and Machine Intelligence}} \bibinfo{volume}{44}, \bibinfo{number}{6} (\bibinfo{year}{2020}), \bibinfo{pages}{2806--2826}.
\newblock


\bibitem[Orts-Escolano et~al\mbox{.}(2016)]%
        {orts2016holoportation}
\bibfield{author}{\bibinfo{person}{Sergio Orts-Escolano}, \bibinfo{person}{Christoph Rhemann}, \bibinfo{person}{Sean Fanello}, \bibinfo{person}{Wayne Chang}, \bibinfo{person}{Adarsh Kowdle}, \bibinfo{person}{Yury Degtyarev}, \bibinfo{person}{David Kim}, \bibinfo{person}{Philip~L Davidson}, \bibinfo{person}{Sameh Khamis}, \bibinfo{person}{Mingsong Dou}, {et~al\mbox{.}}} \bibinfo{year}{2016}\natexlab{}.
\newblock \showarticletitle{Holoportation: Virtual 3d teleportation in real-time}. In \bibinfo{booktitle}{\emph{Proceedings of the 29th annual symposium on user interface software and technology}}. \bibinfo{pages}{741--754}.
\newblock


\bibitem[Park et~al\mbox{.}(2021a)]%
        {park2021nerfies}
\bibfield{author}{\bibinfo{person}{Keunhong Park}, \bibinfo{person}{Utkarsh Sinha}, \bibinfo{person}{Jonathan~T Barron}, \bibinfo{person}{Sofien Bouaziz}, \bibinfo{person}{Dan~B Goldman}, \bibinfo{person}{Steven~M Seitz}, {and} \bibinfo{person}{Ricardo Martin-Brualla}.} \bibinfo{year}{2021}\natexlab{a}.
\newblock \showarticletitle{Nerfies: Deformable neural radiance fields}. In \bibinfo{booktitle}{\emph{Proceedings of the IEEE/CVF International Conference on Computer Vision}}. \bibinfo{pages}{5865--5874}.
\newblock


\bibitem[Park et~al\mbox{.}(2021b)]%
        {Hyper-NeRF}
\bibfield{author}{\bibinfo{person}{Keunhong Park}, \bibinfo{person}{Utkarsh Sinha}, \bibinfo{person}{Peter Hedman}, \bibinfo{person}{Jonathan~T Barron}, \bibinfo{person}{Sofien Bouaziz}, \bibinfo{person}{Dan~B Goldman}, \bibinfo{person}{Ricardo Martin-Brualla}, {and} \bibinfo{person}{Steven~M Seitz}.} \bibinfo{year}{2021}\natexlab{b}.
\newblock \showarticletitle{Hypernerf: A higher-dimensional representation for topologically varying neural radiance fields}.
\newblock \bibinfo{journal}{\emph{arXiv preprint arXiv:2106.13228}} (\bibinfo{year}{2021}).
\newblock


\bibitem[Peng et~al\mbox{.}(2021a)]%
        {peng2021animatable}
\bibfield{author}{\bibinfo{person}{Sida Peng}, \bibinfo{person}{Junting Dong}, \bibinfo{person}{Qianqian Wang}, \bibinfo{person}{Shangzhan Zhang}, \bibinfo{person}{Qing Shuai}, \bibinfo{person}{Xiaowei Zhou}, {and} \bibinfo{person}{Hujun Bao}.} \bibinfo{year}{2021}\natexlab{a}.
\newblock \showarticletitle{Animatable neural radiance fields for modeling dynamic human bodies}. In \bibinfo{booktitle}{\emph{Proceedings of the IEEE/CVF International Conference on Computer Vision}}. \bibinfo{pages}{14314--14323}.
\newblock


\bibitem[Peng et~al\mbox{.}(2023)]%
        {peng2023representing}
\bibfield{author}{\bibinfo{person}{Sida Peng}, \bibinfo{person}{Yunzhi Yan}, \bibinfo{person}{Qing Shuai}, \bibinfo{person}{Hujun Bao}, {and} \bibinfo{person}{Xiaowei Zhou}.} \bibinfo{year}{2023}\natexlab{}.
\newblock \showarticletitle{Representing Volumetric Videos as Dynamic MLP Maps}. In \bibinfo{booktitle}{\emph{Proceedings of the IEEE/CVF Conference on Computer Vision and Pattern Recognition}}. \bibinfo{pages}{4252--4262}.
\newblock


\bibitem[Peng et~al\mbox{.}(2021b)]%
        {peng2021neural}
\bibfield{author}{\bibinfo{person}{Sida Peng}, \bibinfo{person}{Yuanqing Zhang}, \bibinfo{person}{Yinghao Xu}, \bibinfo{person}{Qianqian Wang}, \bibinfo{person}{Qing Shuai}, \bibinfo{person}{Hujun Bao}, {and} \bibinfo{person}{Xiaowei Zhou}.} \bibinfo{year}{2021}\natexlab{b}.
\newblock \showarticletitle{Neural body: Implicit neural representations with structured latent codes for novel view synthesis of dynamic humans}. In \bibinfo{booktitle}{\emph{Proceedings of the IEEE/CVF Conference on Computer Vision and Pattern Recognition}}. \bibinfo{pages}{9054--9063}.
\newblock


\bibitem[Petrovich et~al\mbox{.}(2021)]%
        {petrovich2021action}
\bibfield{author}{\bibinfo{person}{Mathis Petrovich}, \bibinfo{person}{Michael~J Black}, {and} \bibinfo{person}{G{\"u}l Varol}.} \bibinfo{year}{2021}\natexlab{}.
\newblock \showarticletitle{Action-conditioned 3D human motion synthesis with transformer VAE}. In \bibinfo{booktitle}{\emph{Proceedings of the IEEE/CVF International Conference on Computer Vision}}. \bibinfo{pages}{10985--10995}.
\newblock


\bibitem[Pumarola et~al\mbox{.}(2021a)]%
        {D-NeRF}
\bibfield{author}{\bibinfo{person}{Albert Pumarola}, \bibinfo{person}{Enric Corona}, \bibinfo{person}{Gerard Pons-Moll}, {and} \bibinfo{person}{Francesc Moreno-Noguer}.} \bibinfo{year}{2021}\natexlab{a}.
\newblock \showarticletitle{D-nerf: Neural radiance fields for dynamic scenes}. In \bibinfo{booktitle}{\emph{Proceedings of the IEEE/CVF Conference on Computer Vision and Pattern Recognition}}. \bibinfo{pages}{10318--10327}.
\newblock


\bibitem[Pumarola et~al\mbox{.}(2021b)]%
        {pumarola2021d}
\bibfield{author}{\bibinfo{person}{Albert Pumarola}, \bibinfo{person}{Enric Corona}, \bibinfo{person}{Gerard Pons-Moll}, {and} \bibinfo{person}{Francesc Moreno-Noguer}.} \bibinfo{year}{2021}\natexlab{b}.
\newblock \showarticletitle{D-nerf: Neural radiance fields for dynamic scenes}. In \bibinfo{booktitle}{\emph{Proceedings of the IEEE/CVF Conference on Computer Vision and Pattern Recognition}}. \bibinfo{pages}{10318--10327}.
\newblock


\bibitem[Shao et~al\mbox{.}(2023)]%
        {shao2023tensor4d}
\bibfield{author}{\bibinfo{person}{Ruizhi Shao}, \bibinfo{person}{Zerong Zheng}, \bibinfo{person}{Hanzhang Tu}, \bibinfo{person}{Boning Liu}, \bibinfo{person}{Hongwen Zhang}, {and} \bibinfo{person}{Yebin Liu}.} \bibinfo{year}{2023}\natexlab{}.
\newblock \showarticletitle{Tensor4d: Efficient neural 4d decomposition for high-fidelity dynamic reconstruction and rendering}. In \bibinfo{booktitle}{\emph{Proceedings of the IEEE/CVF Conference on Computer Vision and Pattern Recognition}}. \bibinfo{pages}{16632--16642}.
\newblock


\bibitem[Sofianos et~al\mbox{.}(2021)]%
        {sofianos2021space}
\bibfield{author}{\bibinfo{person}{Theodoros Sofianos}, \bibinfo{person}{Alessio Sampieri}, \bibinfo{person}{Luca Franco}, {and} \bibinfo{person}{Fabio Galasso}.} \bibinfo{year}{2021}\natexlab{}.
\newblock \showarticletitle{Space-time-separable graph convolutional network for pose forecasting}. In \bibinfo{booktitle}{\emph{Proceedings of the IEEE/CVF International Conference on Computer Vision}}. \bibinfo{pages}{11209--11218}.
\newblock


\bibitem[Song et~al\mbox{.}(2023)]%
        {song2023nerfplayer}
\bibfield{author}{\bibinfo{person}{Liangchen Song}, \bibinfo{person}{Anpei Chen}, \bibinfo{person}{Zhong Li}, \bibinfo{person}{Zhang Chen}, \bibinfo{person}{Lele Chen}, \bibinfo{person}{Junsong Yuan}, \bibinfo{person}{Yi Xu}, {and} \bibinfo{person}{Andreas Geiger}.} \bibinfo{year}{2023}\natexlab{}.
\newblock \showarticletitle{Nerfplayer: A streamable dynamic scene representation with decomposed neural radiance fields}.
\newblock \bibinfo{journal}{\emph{IEEE Transactions on Visualization and Computer Graphics}} \bibinfo{volume}{29}, \bibinfo{number}{5} (\bibinfo{year}{2023}), \bibinfo{pages}{2732--2742}.
\newblock


\bibitem[Tretschk et~al\mbox{.}(2021)]%
        {tretschk2021non}
\bibfield{author}{\bibinfo{person}{Edgar Tretschk}, \bibinfo{person}{Ayush Tewari}, \bibinfo{person}{Vladislav Golyanik}, \bibinfo{person}{Michael Zollh{\"o}fer}, \bibinfo{person}{Christoph Lassner}, {and} \bibinfo{person}{Christian Theobalt}.} \bibinfo{year}{2021}\natexlab{}.
\newblock \showarticletitle{Non-rigid neural radiance fields: Reconstruction and novel view synthesis of a dynamic scene from monocular video}. In \bibinfo{booktitle}{\emph{Proceedings of the IEEE/CVF International Conference on Computer Vision}}. \bibinfo{pages}{12959--12970}.
\newblock


\bibitem[Villegas et~al\mbox{.}(2018)]%
        {villegas2018hierarchical}
\bibfield{author}{\bibinfo{person}{Ruben Villegas}, \bibinfo{person}{Dumitru Erhan}, \bibinfo{person}{Honglak Lee}, {et~al\mbox{.}}} \bibinfo{year}{2018}\natexlab{}.
\newblock \showarticletitle{Hierarchical long-term video prediction without supervision}. In \bibinfo{booktitle}{\emph{International Conference on Machine Learning}}. PMLR, \bibinfo{pages}{6038--6046}.
\newblock


\bibitem[Vondrick et~al\mbox{.}(2016)]%
        {vondrick2016generating}
\bibfield{author}{\bibinfo{person}{Carl Vondrick}, \bibinfo{person}{Hamed Pirsiavash}, {and} \bibinfo{person}{Antonio Torralba}.} \bibinfo{year}{2016}\natexlab{}.
\newblock \showarticletitle{Generating videos with scene dynamics}.
\newblock \bibinfo{journal}{\emph{Advances in neural information processing systems}}  \bibinfo{volume}{29} (\bibinfo{year}{2016}).
\newblock


\bibitem[Wang et~al\mbox{.}(2018)]%
        {wang2018eidetic}
\bibfield{author}{\bibinfo{person}{Yunbo Wang}, \bibinfo{person}{Lu Jiang}, \bibinfo{person}{Ming-Hsuan Yang}, \bibinfo{person}{Li-Jia Li}, \bibinfo{person}{Mingsheng Long}, {and} \bibinfo{person}{Li Fei-Fei}.} \bibinfo{year}{2018}\natexlab{}.
\newblock \showarticletitle{Eidetic 3D LSTM: A model for video prediction and beyond}. In \bibinfo{booktitle}{\emph{International conference on learning representations}}.
\newblock


\bibitem[Wang et~al\mbox{.}(2023)]%
        {wang2023semantic}
\bibfield{author}{\bibinfo{person}{Zifan Wang}, \bibinfo{person}{Zhuorui Ye}, \bibinfo{person}{Haoran Wu}, \bibinfo{person}{Junyu Chen}, {and} \bibinfo{person}{Li Yi}.} \bibinfo{year}{2023}\natexlab{}.
\newblock \showarticletitle{Semantic Complete Scene Forecasting from a 4D Dynamic Point Cloud Sequence}.
\newblock \bibinfo{journal}{\emph{arXiv preprint arXiv:2312.08054}} (\bibinfo{year}{2023}).
\newblock


\bibitem[Weng et~al\mbox{.}(2022)]%
        {weng2022humannerf}
\bibfield{author}{\bibinfo{person}{Chung-Yi Weng}, \bibinfo{person}{Brian Curless}, \bibinfo{person}{Pratul~P Srinivasan}, \bibinfo{person}{Jonathan~T Barron}, {and} \bibinfo{person}{Ira Kemelmacher-Shlizerman}.} \bibinfo{year}{2022}\natexlab{}.
\newblock \showarticletitle{Humannerf: Free-viewpoint rendering of moving people from monocular video}. In \bibinfo{booktitle}{\emph{Proceedings of the IEEE/CVF conference on computer vision and pattern Recognition}}. \bibinfo{pages}{16210--16220}.
\newblock


\bibitem[Wu et~al\mbox{.}(2023)]%
        {4D-Gaussians}
\bibfield{author}{\bibinfo{person}{Guanjun Wu}, \bibinfo{person}{Taoran Yi}, \bibinfo{person}{Jiemin Fang}, \bibinfo{person}{Lingxi Xie}, \bibinfo{person}{Xiaopeng Zhang}, \bibinfo{person}{Wei Wei}, \bibinfo{person}{Wenyu Liu}, \bibinfo{person}{Qi Tian}, {and} \bibinfo{person}{Xinggang Wang}.} \bibinfo{year}{2023}\natexlab{}.
\newblock \showarticletitle{4d gaussian splatting for real-time dynamic scene rendering}.
\newblock \bibinfo{journal}{\emph{arXiv preprint arXiv:2310.08528}} (\bibinfo{year}{2023}).
\newblock


\bibitem[Xu et~al\mbox{.}(2023)]%
        {xu20234k4d}
\bibfield{author}{\bibinfo{person}{Zhen Xu}, \bibinfo{person}{Sida Peng}, \bibinfo{person}{Haotong Lin}, \bibinfo{person}{Guangzhao He}, \bibinfo{person}{Jiaming Sun}, \bibinfo{person}{Yujun Shen}, \bibinfo{person}{Hujun Bao}, {and} \bibinfo{person}{Xiaowei Zhou}.} \bibinfo{year}{2023}\natexlab{}.
\newblock \showarticletitle{4k4d: Real-time 4d view synthesis at 4k resolution}.
\newblock \bibinfo{journal}{\emph{arXiv preprint arXiv:2310.11448}} (\bibinfo{year}{2023}).
\newblock


\bibitem[Yang et~al\mbox{.}(2023b)]%
        {yang2023diffusion}
\bibfield{author}{\bibinfo{person}{Ruihan Yang}, \bibinfo{person}{Prakhar Srivastava}, {and} \bibinfo{person}{Stephan Mandt}.} \bibinfo{year}{2023}\natexlab{b}.
\newblock \showarticletitle{Diffusion probabilistic modeling for video generation}.
\newblock \bibinfo{journal}{\emph{Entropy}} \bibinfo{volume}{25}, \bibinfo{number}{10} (\bibinfo{year}{2023}), \bibinfo{pages}{1469}.
\newblock


\bibitem[Yang et~al\mbox{.}(2023a)]%
        {Deformable-Gaussian}
\bibfield{author}{\bibinfo{person}{Ziyi Yang}, \bibinfo{person}{Xinyu Gao}, \bibinfo{person}{Wen Zhou}, \bibinfo{person}{Shaohui Jiao}, \bibinfo{person}{Yuqing Zhang}, {and} \bibinfo{person}{Xiaogang Jin}.} \bibinfo{year}{2023}\natexlab{a}.
\newblock \showarticletitle{Deformable 3d gaussians for high-fidelity monocular dynamic scene reconstruction}.
\newblock \bibinfo{journal}{\emph{arXiv preprint arXiv:2309.13101}} (\bibinfo{year}{2023}).
\newblock


\bibitem[Yang et~al\mbox{.}(2023c)]%
        {yang2023real}
\bibfield{author}{\bibinfo{person}{Zeyu Yang}, \bibinfo{person}{Hongye Yang}, \bibinfo{person}{Zijie Pan}, \bibinfo{person}{Xiatian Zhu}, {and} \bibinfo{person}{Li Zhang}.} \bibinfo{year}{2023}\natexlab{c}.
\newblock \showarticletitle{Real-time photorealistic dynamic scene representation and rendering with 4d gaussian splatting}.
\newblock \bibinfo{journal}{\emph{arXiv preprint arXiv:2310.10642}} (\bibinfo{year}{2023}).
\newblock


\bibitem[Yoon et~al\mbox{.}(2020)]%
        {yoon2020novel}
\bibfield{author}{\bibinfo{person}{Jae~Shin Yoon}, \bibinfo{person}{Kihwan Kim}, \bibinfo{person}{Orazio Gallo}, \bibinfo{person}{Hyun~Soo Park}, {and} \bibinfo{person}{Jan Kautz}.} \bibinfo{year}{2020}\natexlab{}.
\newblock \showarticletitle{Novel view synthesis of dynamic scenes with globally coherent depths from a monocular camera}. In \bibinfo{booktitle}{\emph{Proceedings of the IEEE/CVF Conference on Computer Vision and Pattern Recognition}}. \bibinfo{pages}{5336--5345}.
\newblock


\bibitem[Zhang et~al\mbox{.}(2022)]%
        {zhang2022differentiable}
\bibfield{author}{\bibinfo{person}{Qiang Zhang}, \bibinfo{person}{Seung-Hwan Baek}, \bibinfo{person}{Szymon Rusinkiewicz}, {and} \bibinfo{person}{Felix Heide}.} \bibinfo{year}{2022}\natexlab{}.
\newblock \showarticletitle{Differentiable point-based radiance fields for efficient view synthesis}. In \bibinfo{booktitle}{\emph{SIGGRAPH Asia 2022 Conference Papers}}. \bibinfo{pages}{1--12}.
\newblock


\bibitem[Zheng et~al\mbox{.}(2023a)]%
        {Editable-NeRF}
\bibfield{author}{\bibinfo{person}{Chengwei Zheng}, \bibinfo{person}{Wenbin Lin}, {and} \bibinfo{person}{Feng Xu}.} \bibinfo{year}{2023}\natexlab{a}.
\newblock \showarticletitle{Editablenerf: Editing topologically varying neural radiance fields by key points}. In \bibinfo{booktitle}{\emph{Proceedings of the IEEE/CVF Conference on Computer Vision and Pattern Recognition}}. \bibinfo{pages}{8317--8327}.
\newblock


\bibitem[Zheng et~al\mbox{.}(2023b)]%
        {zheng2023pointavatar}
\bibfield{author}{\bibinfo{person}{Yufeng Zheng}, \bibinfo{person}{Wang Yifan}, \bibinfo{person}{Gordon Wetzstein}, \bibinfo{person}{Michael~J Black}, {and} \bibinfo{person}{Otmar Hilliges}.} \bibinfo{year}{2023}\natexlab{b}.
\newblock \showarticletitle{Pointavatar: Deformable point-based head avatars from videos}. In \bibinfo{booktitle}{\emph{Proceedings of the IEEE/CVF Conference on Computer Vision and Pattern Recognition}}. \bibinfo{pages}{21057--21067}.
\newblock


\bibitem[Zielonka et~al\mbox{.}(2023)]%
        {zielonka2023instant}
\bibfield{author}{\bibinfo{person}{Wojciech Zielonka}, \bibinfo{person}{Timo Bolkart}, {and} \bibinfo{person}{Justus Thies}.} \bibinfo{year}{2023}\natexlab{}.
\newblock \showarticletitle{Instant volumetric head avatars}. In \bibinfo{booktitle}{\emph{Proceedings of the IEEE/CVF Conference on Computer Vision and Pattern Recognition}}. \bibinfo{pages}{4574--4584}.
\newblock


\end{thebibliography}


%\appendix
\end{document}